\documentclass[conference]{IEEEtran}

\usepackage{times}

\usepackage{graphicx}
\usepackage{soul}
\usepackage[dvipsnames,table,svgnames]{xcolor}
\usepackage{amssymb}%
\usepackage{amsmath}
\usepackage{booktabs}
\usepackage{pifont}%
\usepackage{multirow}
\usepackage{rotating}
\usepackage{algorithm}
\usepackage{algpseudocode}
\usepackage{xcolor}

\usepackage[numbers]{natbib}
\usepackage{multicol}
\usepackage[bookmarks=true]{hyperref}

\newcommand{\cmark}{\ding{51}}%

\begin{document}

\title{\textbf{AutoMate:} Specialist and Generalist Assembly Policies over Diverse Geometries}

\author{\authorblockN{Bingjie Tang$^{1}$, Iretiayo Akinola$^{2}$, Jie Xu$^{2}$, Bowen Wen$^{2}$,  Ankur Handa$^{2}$,  Karl Van Wyk$^{2}$,  \\ Dieter Fox$^{2,3}$, Gaurav S. Sukhatme$^1$, Fabio Ramos$^{2, 4}$,  Yashraj Narang$^{2}$ }
\authorblockA{ 
$^1$University of Southern California, $^2$NVIDIA Corporation, $^3$University of Washington, $^4$University of Sydney \\
}}

\setcounter{figure}{1}
\makeatletter
\let\@oldmaketitle\@maketitle%
\renewcommand{\@maketitle}{
   \@oldmaketitle%
   \begin{center}
    \centering      
    \includegraphics[width=0.98\textwidth, clip]
    {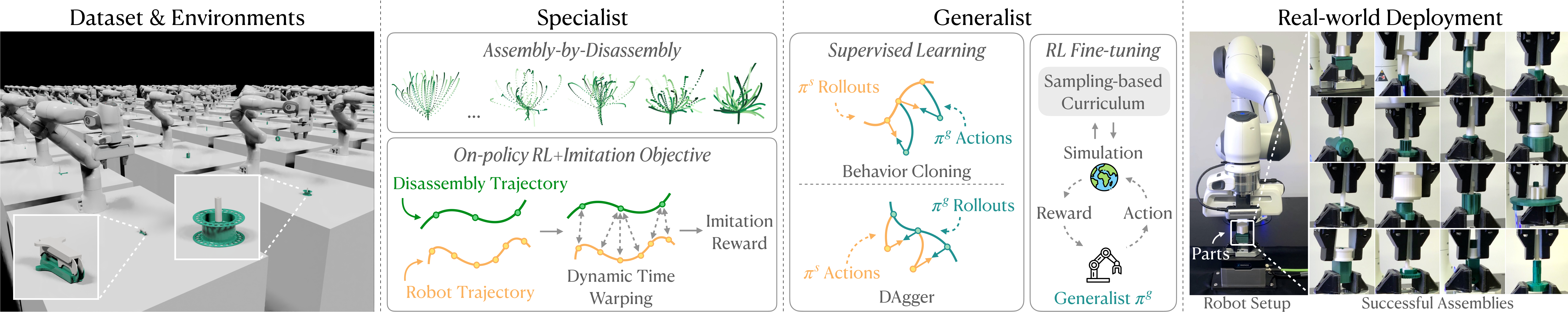}
  \end{center}
  \footnotesize{\textbf{Fig.~\thefigure:\label{fig:teaser}}~ We present A) a dataset of 100 interpenetration-free assemblies that can be simulated in robotics simulators and assembled in the real world, as well as simulation environments for all 100 assemblies in \cite{makoviychuk2021isaac, nvidia2024isaac}; B) specialist (i.e., part-specific) policies trained with a novel approach combining assembly-by-disassembly, RL with imitation, and dynamic time warping, which can solve 80 assemblies with $\approx$80\%+ success rates; C) a generalist (i.e., unified) policy trained with policy distillation and RL fine-tuning, which can solve 20 assemblies with 80\%+ success rates;
    and D) zero-shot sim-to-real transfer of the specialist and generalist policies, including perception-initialized deployments. 
    During evaluations, we execute 5M+ simulated trials and 500 real-world trials.
  }
}
\makeatother 

\maketitle

\begin{abstract}
Robotic assembly for high-mixture settings requires adaptivity to diverse parts and poses, which is an open challenge.
Meanwhile, in other areas of robotics, large models and sim-to-real have led to tremendous progress.
Inspired by such work, we present \textnormal{AutoMate}, a learning framework and system that consists of 4 parts: 1) a dataset of 100 assemblies compatible with simulation and the real world, along with parallelized simulation environments for policy learning, 2) a novel simulation-based approach for learning specialist (i.e., part-specific) policies and generalist (i.e., unified) assembly policies, 3) demonstrations of specialist policies that individually solve 80 assemblies with $\approx$80\%+ success rates in simulation, as well as a generalist policy that jointly solves 20 assemblies with an 80\%+ success rate,
and 4) zero-shot sim-to-real transfer that achieves similar (or better) performance than simulation, including on perception-initialized assembly.\footnote{We use \textit{zero shot} to refer to sim-to-real without a real-world adaptation phase, and \textit{perception-initialized} to refer to assembly that requires perception, grasping, and insertion (rather than starting from a pre-grasped state).}
The key methodological takeaway is that a union of diverse algorithms from manufacturing engineering, character animation, and time-series analysis provides a generic and robust solution for a diverse range of robotic assembly problems.
To our knowledge, \textnormal{AutoMate} provides the first simulation-based framework for learning specialist and generalist policies over a wide range of assemblies,
as well as the first system demonstrating zero-shot sim-to-real transfer over such a range.
For videos and additional details, please see \href{https://bingjietang718.github.io/automate/}{our project website}.

\end{abstract}

\IEEEpeerreviewmaketitle

\section{Introduction}
\label{sec:introduction}

Most objects in home and industrial settings consist of multiple parts that must be assembled \cite{whitney_mechanical_2004}. 
Human workers typically perform assembly; however, in certain industries (e.g., automotive), robotic assembly is prevalent. As industrial robots typically use stiff controllers and perform repetitive motions, robotic assembly requires highly-customized engineering of fixtures, tooling, and waypoints. Nevertheless, in high-mixture settings,
\textit{adaptive} assembly is required \cite{kimble2022performance}, in which robots must assemble parts with diverse geometries and poses. %
Adaptive assembly is non-trivial even for skilled human workers and is a major open challenge in robotics.

Meanwhile, in other areas of robotics, large models and sim-to-real methods have led to tremendous progress. Large language, vision, and visual-language models
have led to vision-language-\textit{action} models with high-level reasoning and multi-task performance \cite{brohan2022rt, brohan2023rt, wayve2023lingo, reed2022generalist}.  Fast simulators (\cite{makoviychuk2021isaac, narang2022factory, todorov2012mujoco}) and domain adaptation (\cite{peng2018sim, ramos2019bayessim, tobin2017domain}) have enabled
robust policies for locomotion \cite{lee2020learning, rudin2021learning} and manipulation \cite{akkaya2019solving, handa2022dextreme}. 
Nevertheless, large models and sim-to-real are extremely
nascent for contact-rich
tasks in industrial settings, including assembly. 
Prior research has often focused on training specialist (i.e., part-specific) policies for 1-5 parts (up to a max of 15-20 parts \cite{spector2021insertionnet, spector2022insertionnet}), and simulation-based development and transfer of generalist (i.e., unified) policies has not been explored. Furthermore, the first zero-shot sim-to-real transfer for perception-initialized assembly has only recently been demonstrated \cite{tang2023industreal}. 

In this context, we present \textbf{AutoMate}, a learning framework and system for solving diverse assembly problems with specialist and generalist policies,
in simulation and with zero-shot transfer to reality
(\textbf{Figure 1}). Our specific contributions are:
\begin{itemize}
    \item \textbf{Dataset and Environments}: We provide a dataset of 100 assemblies based on \cite{tian2022assemble} (but interpenetration-free and with realistic clearances), as well as parallelized simulation environments for all 100 assemblies.
    The datasets and environments provide researchers with a platform for developing policies for a wide range of realistic and realizable (i.e., 3D-printable) assembly problems.
    \item \textbf{Learning Methods}: For training specialists, we propose a novel approach combining 3 distinct algorithms: assembly-by-disassembly, reinforcement learning (RL) with an imitation objective, and dynamic time warping.
    These approaches are a synthesis of diverse algorithms from distinct fields, including manufacturing engineering, character animation, and time-series analysis.
    For training a generalist, we apply geometric encoding, policy distillation, and curriculum-based RL fine-tuning.
    \item \textbf{Specialist and Generalist Policies}: 
    We use our dataset, environments, and methods to learn specialist policies in simulation that can individually solve 80 assemblies with $\approx$80\%+ success rates over 500k trials.
    We also learn a generalist policy in simulation that can jointly solve 20 assemblies with an 80\%+ success rate over 100k trials.
    These results demonstrate that our learning approaches are a generic procedure for solving diverse assembly problems in both a part-specific and unified manner.
    \item \textbf{Sim-to-Real}: We design and demonstrate a real-world system that can deploy our specialist policies in zero-shot with 86.5\% success rates over 20 assemblies and 200 trials, and our generalist policy in zero-shot with an 84.5\% success rate over 20 assemblies and 200 trials. %
    We also execute perception-initialized assembly with 86.0-90.0\% success rates over 5 assemblies and 100 trials.
    These results demonstrate that our learning approaches, when combined with state-of-the-art sim-to-real methods, can produce real-world outcomes that are equivalent to (and sometimes better than) those in simulation.
\end{itemize}

We present the above in \textbf{Sections~\ref{sec:dataset_environments},~\ref{sec:learning_methods},~\ref{sec:specialist_generalist_policies}}, and \textbf{\ref{sec:sim_to_real}}.

To our knowledge, \textbf{AutoMate} provides the first simulation-based framework for learning specialist and generalist policies over a wide range of assemblies,
as well as the first system demonstrating zero-shot sim-to-real transfer over such a range.
Through this work, we aim to gradually build towards the large-model paradigm for industrial robotics, while staying grounded in real-world deployment. We commit to releasing our dataset, environments, and algorithm implementations in the hope of driving forward robotic-assembly research.

\section{Related Works}

Research in robotic assembly has recently experienced significant growth, as described in \cite{narang2022factory, tang2023industreal, tian2022assemble, zhang2023efficient}. 
We focus our review on 1) datasets and benchmarks for assembly of small, realistic parts, 2) our building blocks for learning specialist assembly policies (i.e., assembly-by-disassembly, RL with imitation, and dynamic time warping), 3) our core technique for learning a generalist assembly policy (i.e., policy distillation),
and 4) sim-to-real transfer for assembly.

\subsection{Assembly Datasets and Environments}
\label{sec:related_works_datasets_environments}

There are few existing datasets and environments for assembling small, realistic parts in simulation and the real world.
Simulation efforts include
\cite{willis2022joinable}, a large-scale CAD dataset for realistic assemblies;
\cite{tian2022assemble}, which provides a version suitable for research; 
\cite{narang2022factory}, which provides simulation environments for peg, gear, and connector insertion, and %
\cite{tang2023industreal}, which provides similar environments, as well as a real-world benchmarking kit \cite{tang2023industrealkit}.
The most established real-world effort is
\cite{kimble2020benchmarking, kimble2022performance}, a benchmark for tasks such as connector insertion, pulley alignment, and cable weaving,
as well as a small dataset of CAD models, STL files, images, and point clouds \cite{nist2023moad}.
Finally, other research efforts have provided datasets and environments for additional assembly domains (e.g., \cite{collins2023ramp, heo2023furniturebench, lee2021ikea}). 

Among these efforts, the state-of-the-art may be considered \cite{tian2022assemble} for CAD datasets, \cite{heo2023furniturebench, narang2022factory} for simulation environments, and \cite{kimble2020benchmarking} for real-world benchmarks. Our work draws upon the strengths of each by providing 1) a diverse CAD dataset of 100 interpenetration-free assemblies based on \cite{tian2022assemble}, 2) ready-to-use, parallelized simulation environments for all 100 assemblies in \cite{makoviychuk2021isaac, nvidia2024isaac}, and 3) a real-world benchmarking kit corresponding to the environments.
These components comprise the first unified dataset and environments for sim-to-real transfer for robotic assembly at an appreciable scale.

\subsection{Learning Specialist Assembly Policies}

There are few directly-comparable works for learning specialist policies for assembling a large number of diverse parts (here, 100).
Recent efforts have focused on perception \cite{fu20226d, fu2023lanpose, morgan2021vision, wen2022you} or planning \cite{tian2022assemble, tian2023asap} without learning policies robust to disturbances and noise, or learning policies for a small number of assemblies (1-5), with just a few effort attempting $>$10 assemblies \cite{spector2021insertionnet, spector2022insertionnet, zhao2022offline}.
Thus, we instead review works addressing challenges that we faced when learning specialist policies: 1) generating demonstrations for robotic assembly in simulation, 2) augmenting RL with demonstrations, and 3) selecting relevant demonstrations to use during learning.

For (1), the prevailing approach is assembly-by-disassembly (i.e., generating disassembly sequences/paths and reversing them for use in assembly), which was developed in the context of manufacturing engineering \cite{de1989correct}.
State-of-the-art tree search methods for this process are proposed in \cite{tian2022assemble, tian2023asap}. Importantly, physical laws dictate that only sequences and paths can be reversed 
(rather than velocities and accelerations) \cite{wikipedia2024tsymmetry}.

For (2), there is a diverse set of effective approaches, including bootstrapping RL with behavior-cloned policies \cite{rajeswaran2017learning}, adding demonstrations to the replay buffer for off-policy RL \cite{vecerik2017leveraging}, augmenting the policy gradient for on-policy RL \cite{rajeswaran2017learning}, learning a reward function from demonstrations \cite{ng2000algorithms} or human preferences \cite{christiano2017deep}, and explicitly including an imitation objective in the reward function for on-policy RL \cite{peng2018deepmimic, peng2020learning}, which was proposed in the character-animation literature.

Finally, for (3), the simplest approach is to select the demonstration closest to the initial position of the end effector.
However, as we show, selecting the closest demonstration to the current position at each timestep produces more robust behavior, and selecting the closest demonstration to the \textit{history} of positions (i.e., the end-effector path) is even more effective.
Matching paths of varying lengths and discretizations is a well-known challenge; two state-of-the-art methods are dynamic time warping (DTW) \cite{bellman1959adaptive} and signature transforms \cite{kidger2019deep}, mathematical techniques that were first applied to time-series analysis in speech recognition \cite{sakoe1978dynamic} and finance \cite{gyurko2013extracting}.

Our work combines the strengths of the preceding works by proposing a novel approach combining 3 algorithms: 1) assembly-by-disassembly, 2) RL with an imitation objective, and 3) trajectory matching via DTW. We select (2) for its simplicity and demonstrated effectiveness, and uniquely formulate our imitation objective to imitate paths rather than states or state-action pairs, and we select (3) based on subsequently-described evaluations. This combination enables effective training of specialist policies for $\approx$80\% of our assemblies.

\subsection{Learning Generalist Assembly Policies}

There are few directly-comparable works for simulation-based learning of generalist policies for assembling a large number of parts. We instead contextualize our core technique for learning a generalist policy: policy distillation.

As described in \textbf{Section~\ref{sec:introduction}}, large models have led to tremendous progress in robotics, including multi-task performance.
Whereas such models often call existing skills or train from scratch, we aim for a middle ground by leveraging knowledge from specialist policies via distillation.
In machine learning, \textit{distillation} refers to compressing a neural network by transferring knowledge from a larger \textit{teacher} network to a smaller \textit{student} network \cite{buciluǎ2006model, hinton2015distilling}; in deep RL, \textit{policy} distillation applies this idea to policy networks that map observations to actions \cite{rusu2015policy}.
In robotics, there are two main variations of policy distillation: 
1) \textit{cross-modal distillation}: transferring knowledge from a teacher policy with privileged information (e.g., 6-DOF poses) to a student policy with realistic sensory inputs (e.g., proprioceptive data, RGB images)
\cite{akinola2023tacsl, chen2020learning, chen2022system, lee2020learning, xu2023unidexgrasp}, and 2) \textit{multi-task distillation} (a.k.a., \textit{generalist-specialist learning}): transferring knowledge from multiple task-specific teachers to a single student \cite{ghosh2017divide, jia2022improving, wan2023unidexgrasp++}. Knowledge transfer is typically achieved via behavior cloning (BC) \cite{pomerleau1988alvinn} and/or DAgger \cite{ross2011reduction}.

Inspired by
\cite{wan2023unidexgrasp++}, which focuses on grasping in simulation, our work leverages multi-task distillation to train our generalist policy from our specialist policies via BC, DAgger, and RL fine-tuning.
In contrast, our work avoids the engineering complexity of \cite{wan2023unidexgrasp++}, applies these techniques to robotic assembly, and deploys the learned generalist policy for the first time in the real world (as opposed to simulation only).

\subsection{Sim-to-Real Transfer for Assembly}

There have been numerous sim-to-real efforts for robotic assembly, which have used simulation to enable rapid development, safe policy learning, and scalable experimentation. \cite{tang2023industreal} reviews prior studies in detail and notes that most involve large parts or clearances, use specialized fixtures or adapters, require human demonstrations,
and/or require real-world policy adaptation. We briefly review more recent efforts.

\cite{tang2023industreal} demonstrates zero-shot sim-to-real for perception-initialized assembly on 9 tasks derived from \cite{kimble2020benchmarking}.
They propose algorithms to overcome sim-to-real gap, including SAPU (penalizing interpenetration error during training) and PLAI (integrating actions to reduce steady-state error during deployment).
\cite{zhang2023efficient} demonstrates sim-to-real with pregrasped parts and a real-world adaptation phase for 6 assemblies.
They learn velocity targets and admittance gains for motion primitives in simulation and optimize gains online during deployment.
\cite{tian2023asap} demonstrates sim-to-real for one 5-part assembly.
They develop sequence and path-planning algorithms and execute the demonstration with precisely-fixtured parts.
\cite{zhang2023bridging} demonstrates zero-shot sim-to-real with pregrasped parts for 6 assemblies.
They train a policy in simulation to collect a dataset, which is used to train a 
planner and gain tuner via supervised learning; the planner and tuner are deployed in the real world.

Our work primarily applies the sim-to-real methods described and implemented in \cite{tang2023industreal, tang2023industreallib} and requires no human demonstrations or policy adaptation.
However, we substantially reduce the remaining human effort required by allowing plugs to be placed haphazardly on a platform or in the robot gripper, using a second gripper to grasp the sockets (rather than bolting them to a flat surface), optimizing grasp poses (rather than manually specifying grasp heights), performing 6D pose estimation (rather than detection) during perception-initialized deployments, and using identical control gains and action scales (rather than task-specific scales) for all assemblies.

\section{Problem Description}
\label{sec:problem_description}

Our fundamental task is to use off-the-shelf, research-grade robot hardware
to assemble a wide range of assemblies. Unlike most prior efforts, the assemblies consist of small parts with diverse geometries, 
the parts are initialized with appreciable 6-DOF pose randomization, 
no part-specific adapters or fixtures are leveraged, and no force-torque sensor is used.

Specifically, our experimental setup consists of 1) a Franka Panda robot with shore 30A finger pads mounted to a tabletop, 2) a wrist-mounted Intel RealSense D435 RGB-D camera, 3) a 3D-printed plug and socket\footnote{We use \textit{plug} to refer to a part that must be inserted, and \textit{socket} to refer to a part that mates with the plug.} with 0.5-1.0 mm diametral clearances,
and 4) a Schunk EGK40 gripper with shore 40A gripper pads mounted to the tabletop (akin to \cite{holladay2022robust}). At the beginning of each experiment, the plug is haphazardly pressed into a foam block or placed in the robot gripper, and the socket is haphazardly placed in the Schunk gripper\footnote{Assembly tasks performed by humans typically require two hands, one for manipulation and the other for stabilization; the Schunk gripper allows us to stabilize the socket without incurring the cost of a second robot arm.} (\textbf{Figure~\ref{fig:real-experimental-setup}}).

We assume that 1) each assembly consists of 2 parts (thus, free from sequence planning \cite{tian2022assemble}),
2) all parts have a size and initial position and orientation
such that $\geq$1 grasp is feasible and $\geq$1 feasible grasp is sufficient to allow subsequent insertion (i.e., regrasping is not necessary), 3) a mesh file is available for each part,
which is typical for industrial assembly applications, 4) the end effector can perform assembly in an approximately top-down configuration (i.e., within a 30$^{\circ}$ cone), which is also typical for many applications.

For a formal problem statement, see \textbf{Appendix~\ref{sec:appendix_formal_problem_statement}}.

\section{Dataset and Environments}
\label{sec:dataset_environments}

Our first contribution is a dataset of 100 assemblies compatible with both simulation and the real world, as well as parallelized simulation environments for all 100 assemblies. The key takeaway is that researchers can use these datasets and environments to develop policies for a wide range of realistic and realizable (i.e., 3D-printable) assembly problems.

\subsection{Assembly Dataset}

As described in \textbf{Section~\ref{sec:related_works_datasets_environments}}, \cite{willis2022joinable} provides a large-scale CAD dataset of realistic assemblies, and \cite{tian2022assemble} refines the dataset for research.
However, most meshes still have nonzero interpenetration when assembled; thus, they are incompatible with simulators that enforce non-penetration constraints (e.g., \cite{makoviychuk2021isaac}) and are infeasible to assemble in the real world.

We sample 100 assemblies from \cite{tian2022assemble} that consist of 2 parts, are geometrically diverse, have graspable surfaces, require insertion (rather than simply alignment), and can be assembled approximately top-down. 
Most assemblies have 1 axisymmetric part; however, the part frequently has a symmetry-breaking feature. 
We perform several operations on these meshes: scaling, reorientation, translation, depenetration, chamfering (optional), and subdivision; for details, see \textbf{Appendix~\ref{sec:appendix_mesh_preprocessing}}.
Most critical is depenetration, where we 1) place each plug and socket in their assembled configuration, 2) compute the signed distance from each plug vertex to the surface of the socket \cite{macklin2022warp}, and 3) translate each vertex along its closest face normal until achieving a radial clearance of 0.5~mm.

\begin{figure}
    \centering
    \includegraphics[width=.45\textwidth]{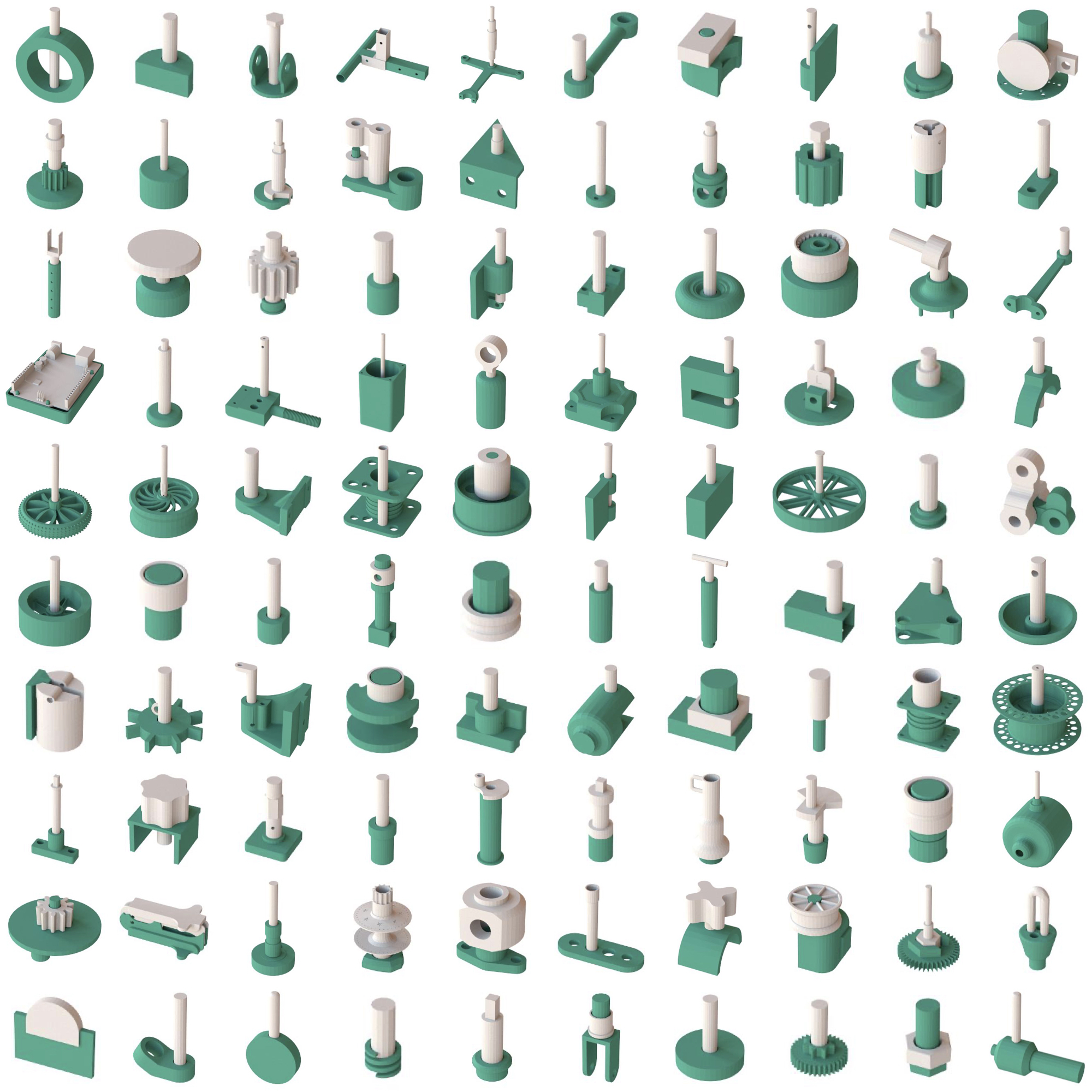}
    \caption{\textbf{Simulation-compatible assembly dataset.} We provide a dataset of 100 assemblies derived from \cite{tian2022assemble}. The assemblies are interpenetration-free, allowing them to be simulated in widely-used robotics simulators.}
    \label{fig:dataset_sim}
\end{figure}

\begin{figure}
\centering\includegraphics[width=.45\textwidth]{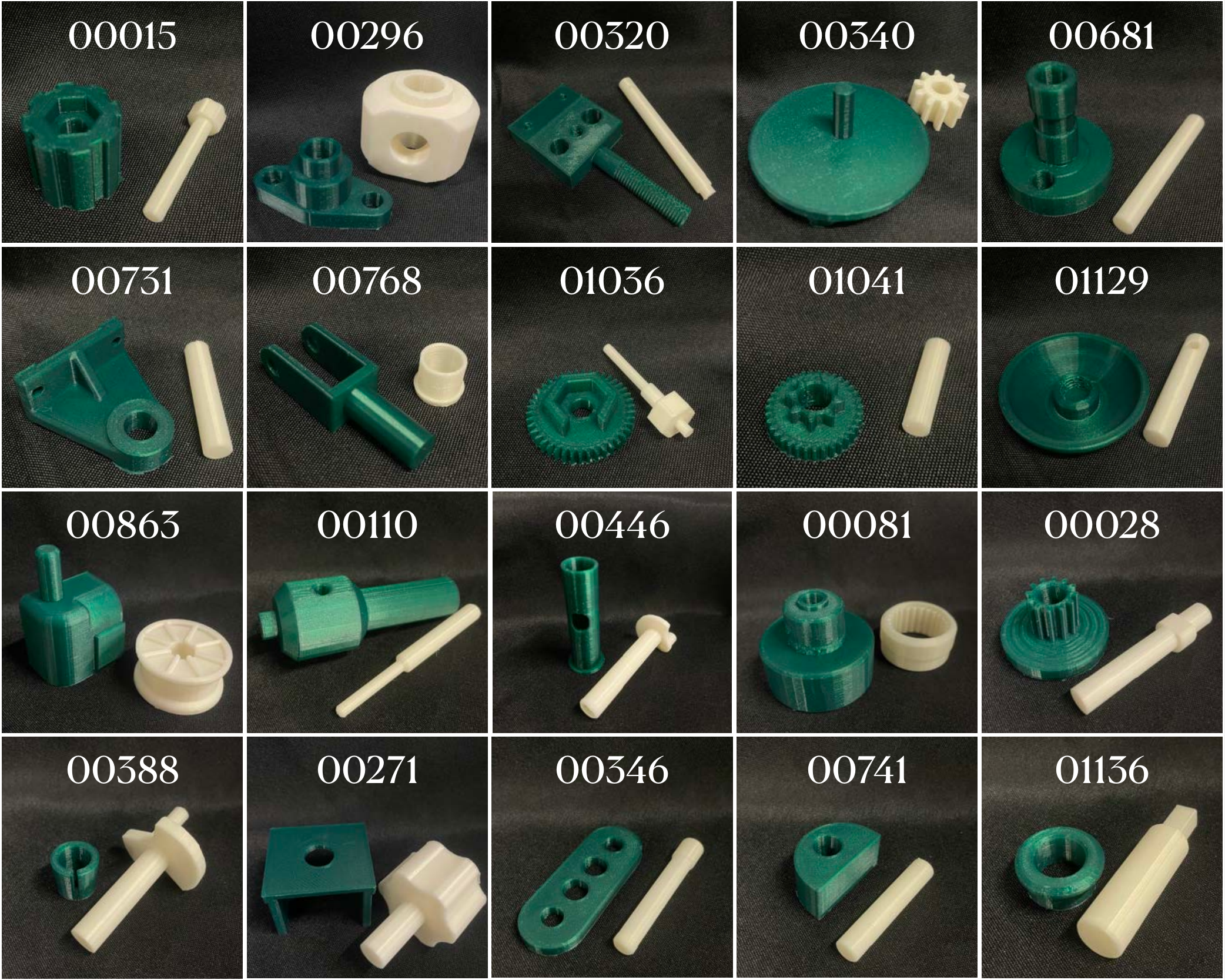}
    \caption{\textbf{Real-world versions of assemblies from our dataset.} We print all 100 assemblies from our dataset in the real world and show 20 assemblies above, with unique IDs listed for later reference.}
    \label{fig:dataset_real}
\end{figure}

The resulting assemblies are all interpenetration-free and have 1~mm of diametral clearance, making them simulation-compatible;
furthermore, they have high triangle density, allowing simulation with fast contact methods that collide meshes against signed distance fields \cite{macklin2020local, narang2022factory} (\textbf{Figure~\ref{fig:dataset_sim}}). 
In addition, they can all be 3D printed and assembled in the real world; due to printer overextrusion, our real-world assemblies have a tighter diametral clearance of 0.5-1.0~mm (\textbf{Figure~\ref{fig:dataset_real}}).

\subsection{Assembly Environments}
\label{subsec:assembly_environments}

We provide ready-to-use, parallelized simulation environments for all 100 assemblies in the dataset in a robotics simulator \cite{makoviychuk2021isaac, nvidia2024isaac}; these environments can be used for arbitrary purposes, including training RL or imitation-learning policies (\textbf{Figure 1 A}).
By default, 
each environment contains a Franka robot, a plug, and a socket.
In the initial state, the robot and socket states are randomized and the plug is randomly initialized in the robot gripper (\textbf{Table~\ref{tab:sim-randomization-ranges}}); in the goal state, each plug is inserted into its corresponding socket.
Initial states can be arbitrarily modified for custom applications.

We also provide utility functions that implement key algorithms used in the work (e.g., trajectory matching based on dynamic time warping).
The simulation environments have been stress-tested by our own research; we aim for them to enable others to train their own specialist and generalist assembly policies and benchmark their results.

\section{Learning Methods}
\label{sec:learning_methods}

Our second contribution is a set of methods for learning specialist and generalist policies over our assembly dataset.
For specialist policies, we find that RL alone is ineffective; thus, we guide RL with imitation learning.
We face 3 challenges: 1) generating demonstrations for assembly, 2) augmenting RL with demonstrations, and 3) selecting demonstrations to use during learning.
To address these challenges, we propose a novel approach combining assembly-by-disassembly, RL with an imitation objective, and trajectory matching via dynamic time warping and signature transforms.
We describe these building blocks in \textbf{Sections~\ref{subsec:assembly-by-disassembly}, \ref{subsec:rl_with_imitation}, and \ref{subsec:trajectory_matching}}, respectively. 

For generalist policies, we face 3 additional challenges: 1) representing assembly geometry to the generalist network, 2) distilling knowledge from the specialists to the generalists, and 3) improving substandard performance. To address these challenges, we apply a combination of geometric encoding, policy distillation, and curriculum-based RL fine-tuning.
We describe these components jointly in \textbf{Section~\ref{subsec:generalist_learning}}.
To see the results of these methods, skip to \textbf{Section~\ref{sec:specialist_generalist_policies}}.

\subsection{Specialist Learning: Assembly-by-Disassembly}
\label{subsec:assembly-by-disassembly}

When training specialists, our first challenge is to generate demonstrations for assembly. Collecting human demonstrations for assembly in simulation is challenging, requiring skilled operators and advanced teleoperation interfaces \cite{mandlekar2023mimicgen}. However, using motion planners is also difficult, as the kinematics of assembly are a narrow passage problem \cite{sun2005narrow}. Inspired by \cite{de1989correct, tian2022assemble}, we instead generate demonstration paths for 
disassembly, which we reverse to generate paths for assembly. 

Specifically, we first perform kinematics- and geometry-based grasp sampling based on \cite{eppner2021acronym}; for details, see \textbf{Figure~\ref{fig:grasp-sampling}} and \textbf{Appendix~\ref{sec:appendix_grasp_optimization}}.
For each assembly, we initialize the plug and socket meshes in their assembled state, sample grasps along the surface of the plug, reject the samples if they violate kinematic or manipulability constraints, and repeat the process until generating 100 grasp candidates.

\begin{figure}
    \centering
    \includegraphics[width=.48\textwidth]{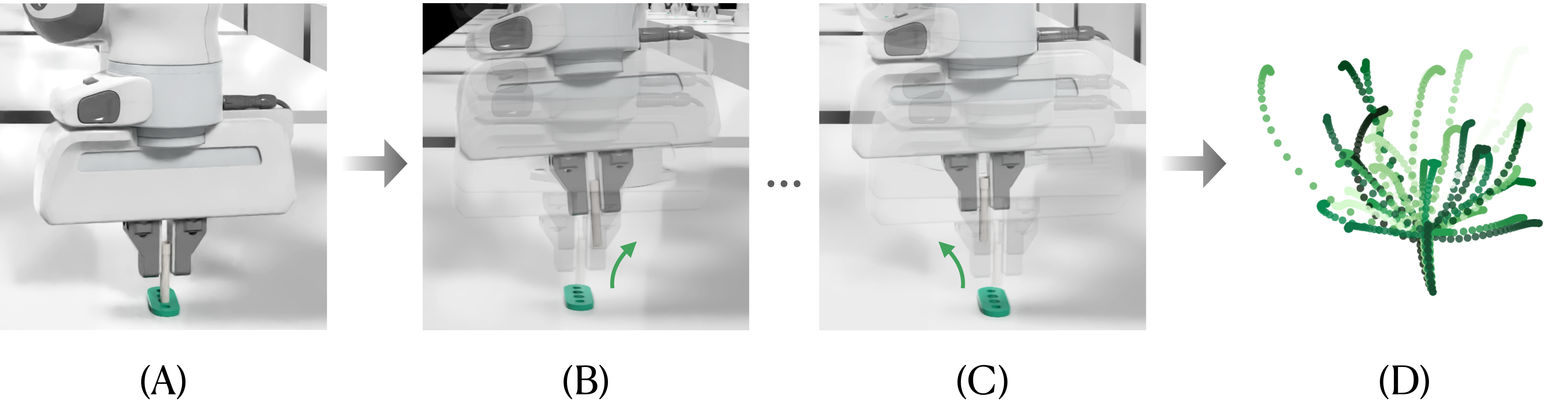}
    \caption{\textbf{Simulation-based generation of disassembly paths.} For each assembly, we generate disassembly paths by A) executing a grasp from a grasp optimization procedure, B) using a low-level controller to lift the plug from the socket and move to a randomized pose, and C) repeating the process for additional poses, until D) collecting 100 successful disassembly paths.}
    \label{fig:disassembly-traj-generation}
\end{figure}

Next, we implement physics-based grasp evaluation. %
For each grasp, we randomize the pose of the assembly (\textbf{Table~\ref{tab:sim-randomization-ranges}}), command the robot to execute the grasp using a low-level controller, and lift the plug to a random pose (\textbf{Figure~\ref{fig:disassembly-traj-generation}}).
If the plug remains in the gripper after the procedure, the grasp is successful.
We repeat the process for 1000 trials and compute success rates for all 100 grasps; the highest-performing grasp for the given assembly is the one with the highest success rate.

For each assembly, now only using the highest-performing grasp, we repeat the evaluation procedure.
For each successful trial, we consider the trajectory of the end effector as a disassembly demonstration $D_i$.
In general, $D_i$ consists of states $x = [x_i^1, x_i^2, ..., x_i^{N_i}]$, where $N$ is the number of states; each state $x_i^j$ can be defined as $[p_i^j; v_i^j; a_i^j]$, where $p$ is position, $v$ is velocity, and $a$ is acceleration.
We define a \textit{reversed} disassembly demonstration $D_i'$ as simply $x_i' = [x_i^{N_i}, x_i^{N_i-1}, ..., x_i^1]$, which can naively be used as an assembly demonstration.
\footnote{We ensure that the range of final poses of the plug in the disassembly paths (i.e., the initial poses of the plug in the reversed disassembly paths) includes the range of initial poses of the plug in the assembly environments (\textbf{Section~\ref{subsec:assembly_environments})}, providing spatially-relevant paths for subsequent imitation.}
However, the corresponding velocities $v_i' = [v_i^{N_i}, v_i^{N_i-1}, ..., v_i^1]$ and accelerations $a_i' = [a_i^{N_i}, a_i^{N_i-1}, ..., a_i^1]$ are in general non-physical. 
Thus, for each successful trial, we record only the reversed disassembly \textit{path} $p_i' = [p_i^{N_i}, p_i^{N_i-1}, ..., p_i^1]$.
We repeat the procedure until collecting 100 successful reversed disassembly paths (\textbf{Figure~\ref{fig:disassembly-traj-generation}D, Figure~\ref{fig:disassembly-paths}}), which we treat as assembly paths in \textbf{Sections~\ref{subsec:rl_with_imitation} and ~\ref{subsec:trajectory_matching}}.

\subsection{Specialist Learning: RL with Imitation Objective}
\label{subsec:rl_with_imitation}

When training specialists, our second challenge is to augment RL with demonstrations. Before we describe our augmentation approach, we briefly describe our baseline RL approach, which is a reimplementation of \cite{tang2023industreal};
for RL formalism and extended descriptions, see \textbf{Appendix~\ref{sec:appendix_reinforcement_learning}}.

As described in \textbf{Section:\ref{subsec:assembly_environments}}, the environments are initialized with a robot, a plug with randomized pose in the robot gripper, and a socket with randomized pose above a tabletop (\textbf{Table~\ref{tab:sim-randomization-ranges}}). Ultimately, the robot must learn a policy that allows it to assemble the plug and socket while being robust to initial randomization and control/perception error.

We formulate the robotic assembly problem as a Markov decision process (MDP), where the objective is to learn a policy
that maximizes the expected sum of discounted rewards (i.e., solves the assembly problem).
We use proximal policy optimization (PPO) \cite{rl-games2022, schulman2017proximal} to learn the policy and an approximation of the value function (hyperparameters in \textbf{Table~\ref{tab:ppo-param}}).

Our observation space
consists of joint angles, the current end-effector pose, and the end-effector goal pose; our input to the critic also includes joint velocities, end-effector linear/angular velocities, and the current plug pose (\textbf{Table~\ref{tab:method-observation}}) \cite{pinto2017asymmetric}).
To model real-world control/perception error, we apply noise to all socket-pose observations (\textbf{Table~\ref{tab:sim-noise}}). Our action space
consists of incremental ($\Delta$) pose targets for a task-space impedance controller. Finally, our reward (without imitation)
consists of terms that 1) penalize distance-to-goal, 2) penalize simulation error, and 3) reward task difficulty at each timestep, as well as a term that rewards task success (\textbf{Appendix~\ref{sec:appendix_reinforcement_learning}}).

Now we describe our augmentation approach. Inspired by \cite{peng2018deepmimic, peng2020learning}, we augment RL with demonstrations 
by directly adding an imitation reward to our reward formulation. Specifically, we define our per-timestep reward as follows:
\begin{equation}
\label{eq:total-reward}
    R_t = \omega^B R_t^B + \omega^I R_t^I
\end{equation}
where $R_t^B$ is the baseline per-timestep reward, described above and in \textbf{Appendix~\ref{sec:appendix_reinforcement_learning}}; $R_t^I$ is an imitation-based per-timestep reward that encourages the agent to mimic demonstrations; and $\omega^B$ and $\omega^I$ are weighting hyperparameters.\footnote{We set $\omega^B$ and $\omega^I$ simply so that the baseline and imitation terms fall within the same order of magnitude.}

Following \cite{peng2018deepmimic}, we define $R_t^I$ as the maximum per-timestep reward over all demonstrations for the given assembly (i.e., the reversed disassembly paths from \textbf{Section~\ref{subsec:assembly-by-disassembly}}). Specifically,
\begin{equation}
\label{eq:avoid-overfitting}
    R_t^I = \max_{i=1,...,M} R^{I_i}_t
\end{equation}
where $M$ is the number of demonstrations.
Unlike \cite{peng2018deepmimic, peng2020learning}, we apply the augmentation approach to contact-rich manipulation rather than locomotion.
We define $R^{I_i}_t$ in \textbf{Section~\ref{subsec:trajectory_matching}}.

\subsection{Specialist Learning: Trajectory Matching via Dynamic Time Warping and Signature Transforms}
\label{subsec:trajectory_matching}

When training specialists, our third and final challenge is to select demonstrations to use during learning.
Specifically, for a given assembly, we must define a reward $R_t^{I_j}$ that quantifies the instantaneous value of imitating any reversed disassembly path $p_i'$, after which we can use \textbf{Equation~\ref{eq:avoid-overfitting}}.

Intuitively, we can define $R_t^{I_i}$ as the distance between the assembly path the robot has already traversed during an RL episode, and the reversed disassembly path $p_i'$ under consideration.
However, like most simulators, ours has a fixed $\Delta t$; thus, for a given path, the arc length between consecutive points is a function of the instantaneous velocity. In general, the path the robot has already traversed has a disparate sequence of velocities compared to any reversed path $p_i'$, resulting in disparate spatial discretizations. We thus seek a distance metric between paths that is insensitive to speed or sampling rate.

We explore 2 powerful methods for computing such a metric, dynamic time warping (\textbf{DTW}) and \textbf{signature transforms}. DTW is a dynamic programming algorithm for quantifying the difference between time series \cite{sakoe1978dynamic}.
Given two sequences $a = [a^1, a^2, ..., a^P]$ and $b = [b^1, b^2, ..., b^Q]$, DTW matches each $a^i$ to one or more $b^j$ and vice versa. The matching process minimizes a cost $C(a, b)$ defined as the sum of Euclidean distances between each $a^i$ and its match(es) from $b$; furthermore, the process satisfies constraints that 1) $a^1$ must match with at least $b^1$ (i.e., first points aligned), 2) $a^P$ must match with at least $b^Q$ (i.e., last points aligned), and 3) all matches must be monotonic (i.e., if $a^i$ matches with $b^j$, then $a^{i+1}$ cannot match with $b^{j-1}$, nor $a^{i-1}$ with $b^{j+1}$). Ultimately, DTW returns the cost $C^*(a, b)$ of the optimal matches between $a$ and $b$; for intuition and pseudocode, see \textbf{Appendix~\ref{sec:appendix_dynamic_time_warping}}.

When we apply DTW, at each timestep $t$, we first extract the path $p_e(t, N)$ of the end effector over a window of length $N$=10 steps, $[p_e^{t-(N-1)}, p_e^{t-(N-2)}, ..., p_e^t]$.
Then, for each reversed disassembly path $p'_i$, we find the closest subsequent points on $p'_i$ from the first point $p_e^{t-(N-1)}$ and current point $p_e^t$ on the windowed end-effector path, and we extract the segment of $p'_i$ between
those 2 closest points. We then use DTW to compute the minimum cost $C^*(p_e(t, N), p_i')$ between the windowed end-effector path and the disassembly segment, and we set $R_t^{I_i} = 1-\tanh{C^*(p_e(t, N), p_i')}$.
We repeat this procedure for each $p'_i$ \cite{cuturi2017soft} and then compute $R_t^I$ (\textbf{Equation~\ref{eq:avoid-overfitting}}).

On the other hand, \textbf{signature transforms} represent trajectories as a collection of path integrals called a path signature \cite{barcelos2023path, chen1958integration, kidger2019deep, lyons1998differential}, which can also quantify distances between paths. 
In our context, given a 3-dimensional path $p(t)_{a,b} = (x_1(t), x_2(t), x_3(t))_{a,b}$, where $x_1(t)$, $x_2(t)$, and $x_3(t)$ represent $x$, $y$, and $z$ coordinates for $t \in [a, b]$, the path signature is a tensor of all possible path integrals between the coordinates. The \textit{first level} of the path signature is
\begin{equation}
    S_1(x_i(t))_{a,t} = \int_a^t dx_i(t) = x_i(t) - x_i(a)
\end{equation}
where $i=1, 2, 3$ (i.e., 3 total integrals), and the \textit{second level} of the path signature is
\begin{equation}
    S_2(x_i(t), x_j(t))_{a,t} = \int_a^t S_1(x_i(t))_{a,t} dx_j(t)
\end{equation}
where $i=1, 2, 3$ and $j=1, 2, 3$ (i.e., 9 total path integrals). Further levels of the path signature can be derived in similar fashion.
Finally, the full path signature is
\begin{equation}
    S(p(t))_{a,b} = (1, S_1(x_i(t)_{a,b}), S_2(x_i(t), x_j(t))_{a,b}, ...)
\end{equation}
where all indices iterate over $1, 2, 3$.\footnote{As our data consists of discrete time series, we technically use the discrete-time form of path signatures (\textbf{Appendix~\ref{sec:appendix_signature_transforms}}).} The signature transform is simply the functional $T(p(t))_{a,b} : p(t)_{a,b} \rightarrow S((p(t))_{a,b}$ that takes a path as input and outputs the path signature. Path signatures inherit translation and reparameterization invariance from path integrals; as desired, these properties mitigate discretization sensitivity.
For details, see \textbf{Appendix~\ref{sec:appendix_signature_transforms}}.

When we apply the signature transform, at each timestep $t$, we consider the full path $ p_e(T)_{0,t}$ of the end effector from the beginning of the episode. Then, for each reversed disassembly path $p_i'$, we find the closest point on $p_i'$ from the current point $p_e(t)$ on the end-effector path and extract the segment of $p_i'$ between the start and the closest point. We then 
compute the path signatures $S(p_e(T))_{0,t}$ and $S(p_i')$ of the end-effector path and disassembly path segment \cite{kidger2021signatory}, respectively, and compute the cost $C(S(p_e(T))_{0,t}, S(p_i'))$ between signatures as
\begin{equation}
    C\big(S(p_e(T))_{0,t}, S(p_i')\big) = \| S(p_e(T))_{0,t} - S(p_i') \|_2.
\end{equation}

Finally, we set $R_t^{I_i} = 1-\tanh{C\big(S(p_e(T))_{0,t}, S(p_i')\big)}$.

We have thus described our specialist learning methods, which consist of assembly-by-disassembly, RL with an imitation objective, and trajectory matching. To see evaluations and results, skip to \textbf{Section~\ref{sec:specialist_generalist_policies}}. Next, we describe our generalist learning methods, which consist of geometric encoding, policy distillation, and curriculum-based RL fine-tuning.

\subsection{Generalist Learning: Geometric Encoding, Policy Distillation, and Curriculum-based RL Fine-tuning}
\label{subsec:generalist_learning}

The first challenge when training a generalist is to represent assembly geometry to the generalist network. For specialist policies, the policy/value networks do not take geometry as input, as it is constant; however, for a generalist policy, the networks must take geometry as input, as assembling a wide range of parts without such knowledge would be infeasible.

Inspired by \cite{wan2023unidexgrasp++}, we learn a latent representation of object geometry prior to learning a generalist policy. Specifically, we sample point clouds from the surfaces of all plug and socket meshes, and we train a PointNet autoencoder \cite{qi2017pointnet} over the point clouds to minimize reconstruction loss;
for details, see \textbf{Appendix~\ref{sec:appendix_point_cloud_autoencoder}}.
We then pass the point clouds into the encoder and store the latent vectors $z_i$. During generalist learning, for a given assembly, we simply concatenate $z_p$ and $z_s$ for the corresponding plug and socket meshes as input to the policy.

The second challenge when training a generalist is to distill knowledge from the specialist policies into a generalist policy. As shown in \textbf{Section~\ref{subsec:generalist_results}}, it is ineffective to train a generalist policy from scratch over a large number of assemblies, and we aim to reuse knowledge from already-trained specialists. Thus, we implement a simple 2-stage policy distillation procedure:
\begin{enumerate}
    \item \textbf{Behavior Cloning (BC)} \cite{pomerleau1988alvinn}: 
    We use standard BC on the specialists.
    Specifically, we execute each specialist policy $\pi_s$ under initial-pose randomization (\textbf{Table~\ref{tab:sim-randomization-ranges}}) and observation noise (\textbf{Table~\ref{tab:sim-noise}}) until completing 5000 successful episodes.
    For each success, we record the state-action pairs as a demonstration $D_i = \{(s_i^1, a_i^1), (s_i^2, a_i^2), ..., (s_i^{N_i}, a_i^{N_i})\}$. 
    We randomly initialize a generalist policy $\pi_g$ and minimize MSE loss $\mathcal{L} = \frac{1}{M} \sum_{i=1}^{M} \sum_{j=1}^{N_i} (a_i^j - \pi_g(s_i^j))^2$ between ground-truth actions in the demonstrations and actions predicted by the generalist  (batch size $M$=128, epochs=1000).
    The output is an initial generalist policy $\pi_g$.
    \item \textbf{DAgger} \cite{ross2011reduction}: 
    We use DAgger to reduce covariate shifts by executing the generalist policy and querying the specialists under the induced state distributions. 
    Specifically, we execute the current $\pi_g$ for $256$ successful or unsuccessful episodes and record the state-action pairs. We 
    minimize MSE loss $\mathcal{L} = \frac{1}{M} \sum_{i=1}^{M} \sum_{j=1}^{N_i} (\pi_g(s_i^j) -\pi_s(s_i^j))^2$ between the actions taken by the generalist and the specialist actions queried at those same states. 
    The output is a refined generalist policy $\pi_g'$.
    
\end{enumerate}

The third challenge when training a generalist is to further improve performance. Although BC and DAgger can produce reasonable policies, BC is limited by dataset size and diversity, and DAgger is limited by specialist performance on states visited by the generalist.
In contrast to the iterative, highly-complex approach of \cite{wan2023unidexgrasp++}, we perform a single RL fine-tuning phase on the generalist that follows the same baseline RL-only approach we evaluate when learning specialists. As we show later (\textbf{Section~\ref{sec:specialist_generalist_policies}}), the curriculum is particularly critical.

Specifically, we initialize our actor with $\pi_g'$ and follow the baseline RL procedure outlined in \textbf{Appendix~\ref{sec:appendix_reinforcement_learning}}. We use a sampling-based curriculum (SBC), where we expose the agent to the full range of initial-pose randomization at the start of the curriculum, but increase the lower bound at each stage; this curriculum outperforms naive implementations for assembly \cite{tang2023industreal}.  More precisely, at each curriculum stage $k={1,...,K}$,
the initial plug height $h_k^{init} \sim U[{h_k^{min}}, h^{max}]$, where $h_k^{min} < h_{max}$, $h_1^{init} \leq h_k^{init} \leq h_K^{init}$, and $h^{max}$ remains constant.

\section{Specialist and Generalist Policies}
\label{sec:specialist_generalist_policies}

Our third contribution is a demonstration that the prior learning methods enable high-performance specialist and generalist policies in simulation. We now present detailed evaluations of our specialist and generalist policies.
The key takeaway is that our learning approaches (\textbf{Section~\ref{sec:learning_methods}}) are a generic and powerful procedure for solving diverse assembly problems in both a part-specific and unified manner.

\begin{figure}[ht]
    \centering
    \includegraphics[width=0.95\columnwidth]{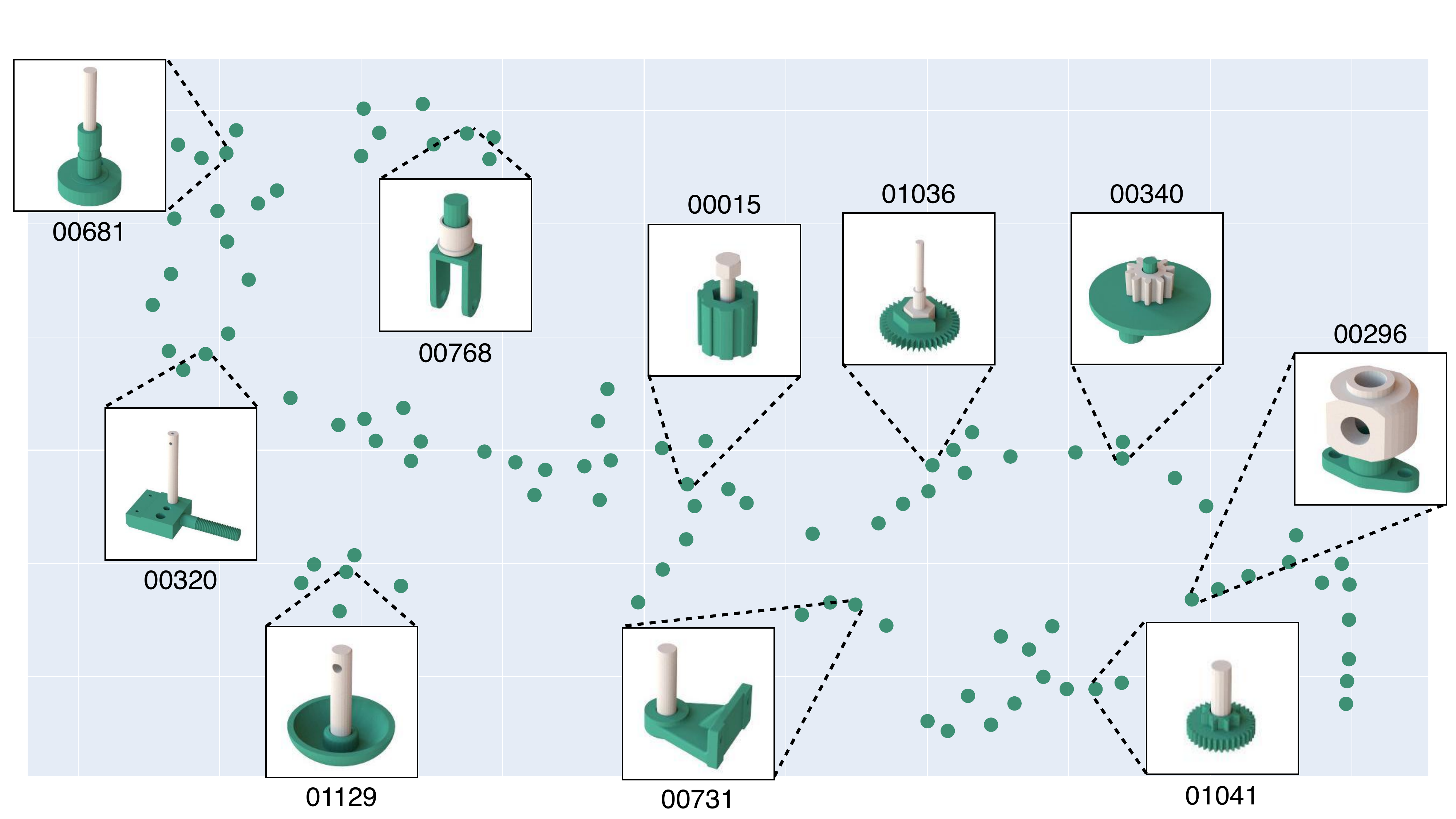}
    \caption{\textbf{t-SNE visualization of geometric representations of 100 assemblies.} We train a PointNet-based autoencoder to learn a latent representation of assembly geometry, and we use t-SNE (with perplexity = 6) to reduce the dimensionality of the latent vectors to 2D. Here we plot the lower-dimensional representations of all 100 assemblies. For visualization, we sample 10 assets that are well distributed across clusters. We also show examples of multiple assemblies sampled from the same cluster in \textbf{Figure~\ref{fig:t-sne_clusters}}.}
    \label{fig:t-sne_selected_assets}
\end{figure}

As a preliminary step to help with exposition, we take the latent vectors of assembly geometry from \textbf{Section~\ref{subsec:generalist_learning}} and use the t-SNE algorithm \cite{hinton2002stochastic} to reduce the dimensionality of the data to 2D (\textbf{Figure~\ref{fig:t-sne_selected_assets}}). We then sample 10 assemblies that are well distributed across the resulting clusters. Although we perform simulation-based evaluations across all 100 assets, we will frequently discuss these 10 assemblies in further detail.

\subsection{Evaluations of Specialist Policies}
\label{subsec:specialist_results}

We now present the results of our specialist policies, which are trained based on our combination of assembly-by-disassembly, RL with an imitation objective, and trajectory matching via dynamic time warping and signature transforms. We perform all evaluations in this section under the maximum bounds for initial-pose randomization (\textbf{Table~\ref{tab:sim-randomization-ranges}}) and observation noise (\textbf{Table~\ref{tab:sim-noise}}) experienced during training.

For specialist policies, our first evaluation question is, \textbf{which trajectory-matching approach is most effective for AutoMate?} We evaluate the following 4 test cases:

\begin{itemize}
    \item \textbf{IndustReal} \cite{tang2023industreal}: This is a state-of-the-art RL-only approach for simulation-based robotic assembly, described in \textbf{Section~\ref{subsec:rl_with_imitation}}. It does not use an imitation objective and thus illustrates results without trajectory matching.
    \item \textbf{State-based Matching}: This is a naive baseline for RL with an imitation objective.
    At each timestep, we calculate the distance from the end effector to every point of every disassembly path, and we compute our imitation reward based on the shortest distance.
    \item \textbf{Dynamic Time Warping (Ours)}: We compute our imitation reward based on the DTW distance (\textbf{Section~\ref{subsec:trajectory_matching}}).
    \item \textbf{Signature Transform (Ours)}: We compute our imitation reward based on a path-signature distance (\textbf{Section~\ref{subsec:trajectory_matching}}).
\end{itemize}

\begin{figure*}[ht]
\centering\includegraphics[width=.98\textwidth]{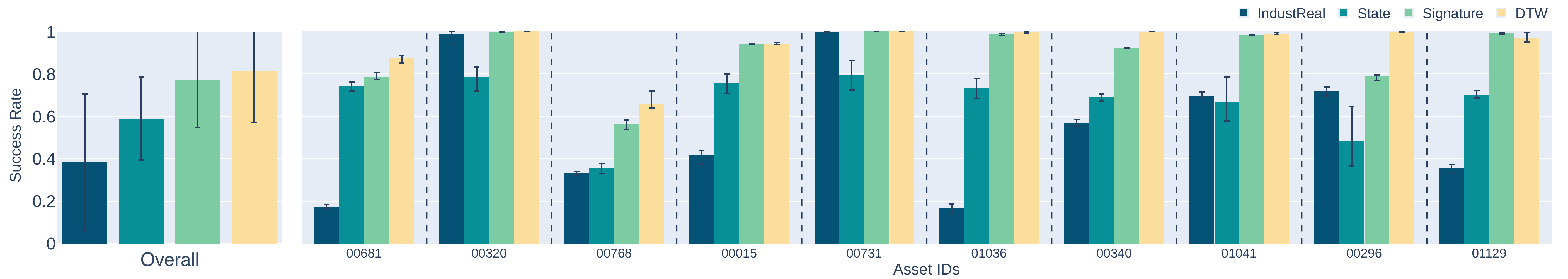}
    \caption{\textbf{Simulation-based evaluation of trajectory-matching approaches for learning specialist policies.} For each of the 100 assemblies, we train a specialist policy with 4 different approaches for matching the current robot path with demonstrations. For each approach, we train 5 random seeds, select the best seed, and evaluate it 5 times over 1000 trials. We illustrate average results over all 100 assemblies, as well as specific results for 10 sampled assemblies (\textbf{Figure~\ref{fig:t-sne_selected_assets}}). \textit{IndustReal} is a state-of-the-art matching-free approach. \textit{State} selects the demonstration containing the closest point to the current robot state. \textit{Signature} selects the demonstration with the minimum signature-transform distance from the robot trajectory. \textit{DTW} selects the demonstration with the minimum dynamic-time-warping distance from the robot trajectory. The \textit{Signature} and \textit{DTW} approaches significantly outperform the others.}
    \label{fig:results_specialist_traj_match}
\end{figure*}

For each of the 4 test cases, for each of the 100 assemblies, we train a specialist policy over 5 random seeds. We select the best seed and evaluate it over 5000 trials, for a total of 2M simulated trials. \textbf{Figure~\ref{fig:results_specialist_traj_match}} shows our results.

IndustReal has the lowest success rates over the 10 selected assemblies, indicating the importance of imitation. State-based matching provides a substantial improvement, but still does not result in high success rates. Dynamic time warping and signature transforms consistently have the highest success rates, with slightly better performance for the former. These results are also reflected over all 100 assemblies, with mean success rates of 38.45 $\pm$ 32.16\% for IndustReal, 59.11 $\pm$ 19.65\% for state-based matching, 81.50 $\pm$ 24.42\% for DTW, and 77.46 $\pm$ 22.61\% for signature transforms.

Our next evaluation question is, \textbf{does AutoMate perform better than naive and state-of-the-art baselines?} We evaluate the following 5 test cases:
\begin{itemize}
    \item \textbf{Go to Goal}: This is a naive baseline for control. We use a task-space impedance controller to move the end effector directly to the goal.
    \item \textbf{Top Down}: This is a naive baseline for control. We use a task-space impedance controller to move the end effector directly above the goal and straight downward.
    \item \textbf{Follow Trajectory}: This is a naive baseline for imitation learning. We select the demonstration containing the closest point to the initial end-effector position, and we use a task-space impedance controller to move the end effector along that fixed path.
    \item \textbf{IndustReal}: This is the state-of-the-art RL-only approach for simulation-based robotic assembly described earlier.
    \item \textbf{AutoMate (Ours)}: We use a combination of assembly-by-disassembly, RL with an imitation objective, and trajectory matching with dynamic time warping.
\end{itemize}

\begin{figure*}
    \centering\includegraphics[width=.98\textwidth]{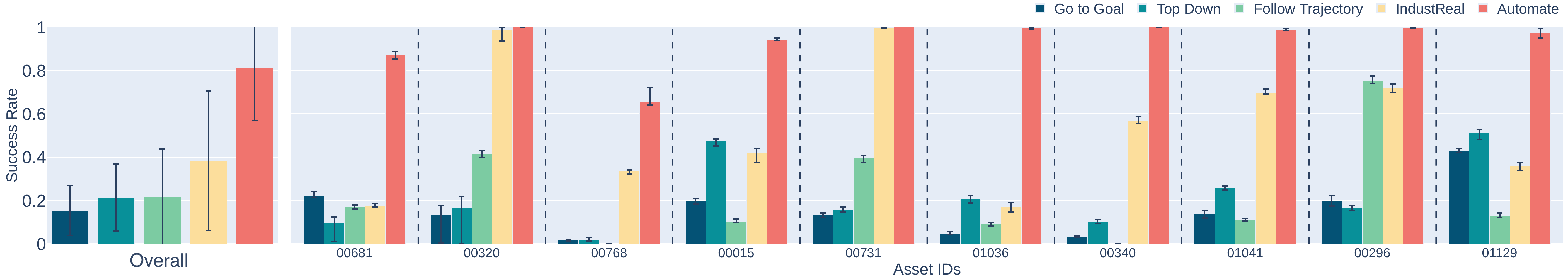}
    \caption{\textbf{Simulation-based evaluation of control schemes for specialist policies.} For each of the 100 assemblies, we build a specialist policy with 5 different control approaches. For each approach, we train 5 random seeds, select the best seed, and evaluate it 5 times over 1000 trials. We illustrate average results over all 100 assemblies, as well as specific results for 10 sampled assemblies (\textbf{Figure~\ref{fig:t-sne_selected_assets}}). \textit{Go to Goal} uses a task-space impedance controller to move directly to the goal. \textit{Top Down} uses the controller to move directly above the goal and then downward. \textit{Follow Trajectory} selects the demonstration containing the point closest to the initial robot state and uses the controller to follow the demonstration path. \textit{IndustReal} is a state-of-the-art, RL-only baseline. \textit{AutoMate} is our proposed strategy combining assembly-by-disassembly, RL with imitation, and dynamic time warping. \textit{Automate} significantly outperforms the others.}
    \label{fig:results_specialist_ctrl_schemes}
\end{figure*}

For each of the 5 test cases, for each of the 100 assemblies, we train a specialist policy over 5 seeds. We select the best seed and evaluate it over 5000 trials, for a total of 2.5M simulated trials. \textbf{Figure~\ref{fig:results_specialist_ctrl_schemes}} shows our results.

Go to Goal, Top Down, and Follow Trajectory have the lowest success rates over the 10 assemblies, as they lack robustness to controller error, observation noise, and disturbances during contact.
IndustReal notably outperforms the 3 naive baselines on several assemblies, demonstrating the benefit of an RL-based approach.
However, AutoMate consistently has the highest success rates. Again, the preceding results are reflected over all 100 assemblies, with mean success rates of $15.44\pm 11.56\%$ for Go To Goal, $21.54\pm 15.49\%$ for Top Down, $21.62\pm 22.32\%$ for Follow Trajectory, $38.45 \pm 32.16\%$ for IndustReal, and $81.50\pm 24.42\%$ for AutoMate.

Our third evaluation question is, \textbf{how does AutoMate perform across all 100 assemblies?} 
Specifically, rather than selecting 10 assemblies or summarizing statistics, we tabulate results over all 100 assemblies. 
\textbf{Appendix~\ref{sec:appendix_specialist_results}} shows our results.
We provide our raw data as supplementary information.

AutoMate shows consistent performance over a wide spectrum of the assemblies.
Specifically, for 80 assemblies, AutoMate has $\approx$80\% (78\%) success rates or higher, and for 55, it has 90\% success rates or higher.
We conclude that AutoMate is a highly-effective strategy for training specialists over diverse geometries. The most challenging assemblies (i.e., bottom 20\%) tend to have small-diameter plugs, sockets with small contact surfaces, or sockets with stair-like internal features.

We conduct additional evaluations of the robustness of our specialist policies to initial-pose randomization in \textbf{Appendix~\ref{sec:appendix_specialist_robustness_initial_pose}} and observation noise in \textbf{Appendix~\ref{sec:appendix_specialist_robustness_obs_noise}}.

\subsection{Evaluation of Generalist Policy}
\label{subsec:generalist_results}

We now present the results of our generalist policies, which are trained based on our combination of behavior cloning, policy distillation, and curriculum-based RL fine-tuning. As with the specialists, we perform all evaluations under the maximum bounds for initial-pose randomization (\textbf{Table~\ref{tab:sim-randomization-ranges}}) and observation noise (\textbf{Table~\ref{tab:sim-noise}}) experienced during training.

Our primary evaluation question is, \textbf{which policy distillation approach is most effective for AutoMate?} We evaluate the following 4 test cases:

\begin{itemize}
    \item \textbf{Behavior Cloning (BC):} We use standard BC to distill the specialist policies to a generalist policy $\pi_g$, as described in \textbf{Section~\ref{subsec:generalist_learning}}.
    \item \textbf{BC + DAgger:} We first use standard BC and then use DAgger to produce a refined generalist policy $\pi_g'$.
    \item \textbf{BC + DAgger + RL fine-tuning:} We first use standard BC, then use DAgger, and finally use RL fine-tuning to produce a generalist policy $\pi_g''$.
    \item \textbf{BC + DAgger + RL fine-tuning with SBC (Ours):} We do the above, but also apply a sampling-based curriculum (SBC) proposed in \cite{tang2023industreal}, where the lower bound of initial-pose randomization increases at each stage, but the upper bound remains fixed (\textbf{Appendix~\ref{sec:appendix_reinforcement_learning}}).
\end{itemize}

\begin{figure*}
\centering\includegraphics[width=.98\textwidth]{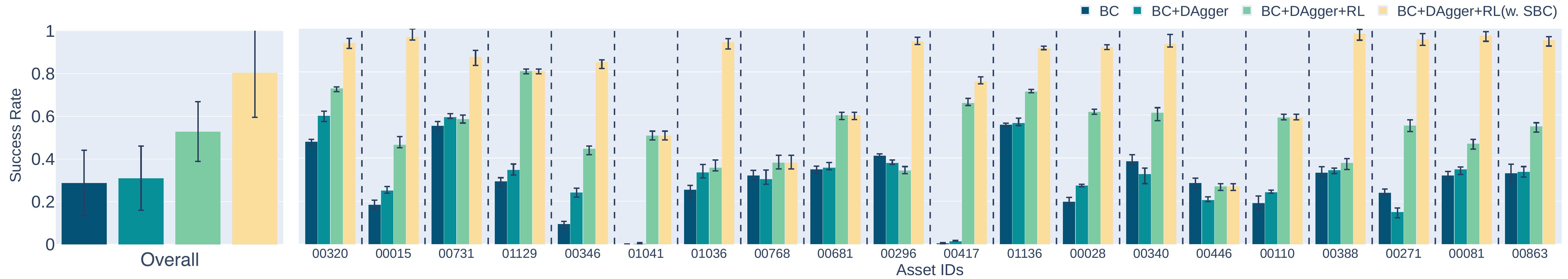}
    \caption{\textbf{Simulation-based evaluation of training approaches for generalist policies.} We train a single generalist policy for 20 assemblies using 4 different approaches. For each approach, we train 5 random seeds, select the best seed, and evaluate it 5 times over 1000 trials. We illustrate average results over the 20 assemblies, as well as specific results for each assembly (asset ID lookup in \textbf{Figure~\ref{fig:asset-lookup}}). \textit{BC} uses behavior cloning to distill the corresponding specialist policies for the 20 assemblies into a generalist policy. \textit{BC + DAgger} follows BC with DAgger iterations to reduce covariate shift. \textit{BC + DAgger + RL} follows \textit{BC + DAgger} with an RL-based fine-tuning stage. \textit{BC + DAgger + RL (w/SBC)} uses a sampling-based curriculum (SBC) for initial-pose randomization \cite{tang2023industreal}, which provided a critical boost in performance; \textit{BC + DAgger + RL (w/SBC)} significantly outperforms the others.}
    \label{fig:results_generalist_distill_methods}
\end{figure*}

For each of the 4 test cases, we train a generalist policy over 20 assemblies; these include the 10 assemblies sampled from t-SNE clusters (\textbf{Figure~\ref{fig:t-sne_selected_assets}}), as well as 10 additional assemblies evenly sampled in t-SNE space.
We evaluate the performance of each test case over 5000 trials per assembly, for a total of 400k trials. 
\textbf{Figure~\ref{fig:results_generalist_distill_methods}} shows our results.

As expected, BC has the lowest success rate over the 20 assemblies (28.84 $\pm $15.23\%), likely due to covariate shift; however, BC + DAgger provides marginal improvement (31.06 $\pm$ 15.06\%).
In contrast, BC + DAgger + RL provides substantial improvement (52.85 $\pm$ 14.01\%), including on assemblies where the prior techniques failed, demonstrating the value of RL-based fine-tuning.
Nevertheless, BC + DAgger + RL with SBC has the highest success rate by far (80.42 $\pm$ 20.93\%) and improves performance on almost every assembly, underscoring curriculum learning during fine-tuning.
As a final sanity check, we train an RL policy with SBC from scratch over the 20 assemblies and measure a low success rate (48.43 $\pm$ 15.28\%).

As a secondary evaluation question, we also ask, what is the scaling law between generalist performance and the number of specialists used in training? For details, see \textbf{Appendix~\ref{sec:appendix_generalist_results}}.

\section{Sim-to-Real Transfer}
\label{sec:sim_to_real}

Our final contribution is sim-to-real transfer of our specialist and generalist policies.
We first describe our real-world system design and then present real-world evaluations and demonstrations of our specialist and generalist policies. We strongly encourage readers to view our supplementary video.
The key takeaway is that our learning approaches, when combined with state-of-the-art sim-to-real transfer and pose estimation methods, can produce real-world outcomes that are equivalent to (and sometimes better than) those in simulation.

\subsection{Real-World System Design}
\label{subsec:system_design}

As first described in \textbf{Section~\ref{sec:problem_description}}, our real-world system consists of a Franka robot with a parallel-jaw gripper, a wrist-mounted RealSense D435 camera, a Schunk EGK40 parallel-jaw gripper mounted to the tabletop, and 3D-printed assemblies from our dataset (\textbf{Figure~\ref{fig:real-experimental-setup}}). 
Our communications framework is closely modeled after \cite{tang2023industreal}; 
however, our perception, grasping, and control procedures differ significantly.

For perception, we aim to accurately estimate plug and socket states while initializing them in a far less-constrained manner.
We use a powerful segmentation tool \cite{kirillov2023segment}, textureless CAD models of our parts, and a state-of-the-art pose estimator \cite{wen2023foundationpose} to estimate the 6-DOF poses of each part from RGB-D images. \textbf{Figure~\ref{fig:perception-pipeline}} shows our pipeline; for details, see \textbf{Appendix~\ref{sec:appendix_perception}}.
For grasping, we aim to avoid manually specifying grasp poses. %
We use our grasp sampling and evaluation procedure described (\textbf{Section~\ref{sec:appendix_grasp_optimization}}) to generate a high-performing grasp for each assembly, and we execute those grasps in the real world.
Finally, we aim to avoid any assembly-specific tuning of controller gains or action scales, and we restrict ourselves to a single set of parameters for all real-world trials.

\begin{figure*}
\centering\includegraphics[width=1.0\textwidth]{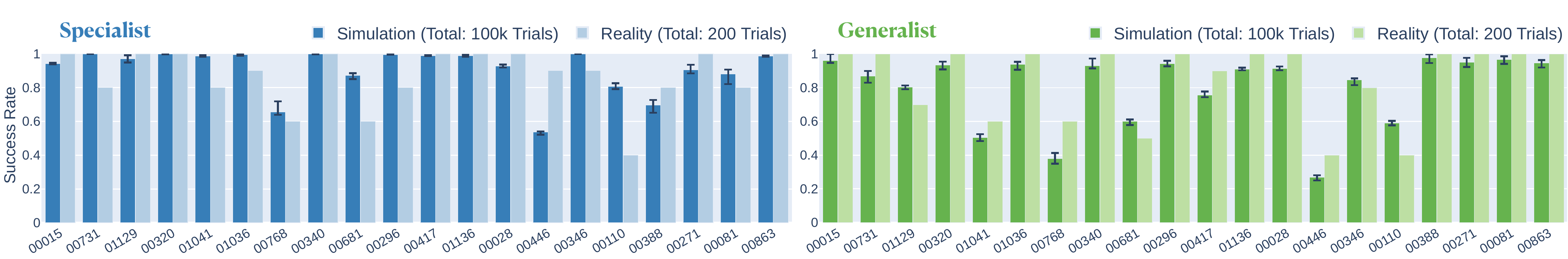}
    \caption{\textbf{Comparison of real-world specialist-policy and generalist-policy success rates with simulated analogues.}  We deploy our specialist policies over 20 assemblies and 200 trials (asset ID lookup in \textbf{Figure~\ref{fig:asset-lookup}}), and we deploy our generalist policy over the same conditions. We compare the results to simulated analogues. Left: For specialists, our success rates in the real world are highly comparable to those in simulation, with a drop of only 4.15\% on average. Right: For the generalist, our success rates in the real world are again highly comparable to simulation, with an \textit{improvement} of 4.08\% on average. %
    Now comparing our real-world specialists to our real-world generalist, we note that generalist performance is within 2.0\% of the specialists on average.}
    \label{fig:eval_sim_real}
\end{figure*}

\subsection{Real-World Policy Evaluations}
\label{subsec:real-world_results}

We now present the results of our specialist and generalist policies in the real world. For these trials, we place the robot in lead-through, manually grasp a plug, and guide it into the socket. We then programmatically lift the plug until free from contact; apply an $xy$ perturbation of $\pm$ 10 mm, $z$ perturbation of 15 $\pm$ 5 mm, and yaw perturbation of $\pm$ 5$^\circ$; apply $x$, $y$, and $z$ observation noise of $\pm2$ mm each; and deploy a policy.\footnote{The manual grasping step is uncontrolled and likely contributes an additional 1-3 mm and 5-10 deg of perturbation.}

Our first evaluation question is, \textbf{do our specialist policies transfer to the real world?} For each of 20 assemblies, we deploy the corresponding specialist policy 10 times, for a total of 200 trials. \textbf{Figure~\ref{fig:eval_sim_real}} (left) shows our results.

The mean success rate in the real world is 86.50 $\pm$ 16.52\%, whereas the success rate in simulation is 90.65 $\pm$ 13.07\%.
Across assemblies, real-world success rates are within close range of simulation; in fact, for 11/20 assemblies, real-world is better.
Such results likely indicate that our simulated training conditions (e.g., initial-pose randomization, observation noise) are sufficiently adverse to train robust and performant policies.
We conduct additional evaluations of the robustness of our policies to initial-pose randomization in \textbf{Appendix~\ref{sec:appendix_sim_to_real_robustness_randomization}}.

\begin{figure*}
    \centering
    \includegraphics[width=1.0\textwidth]{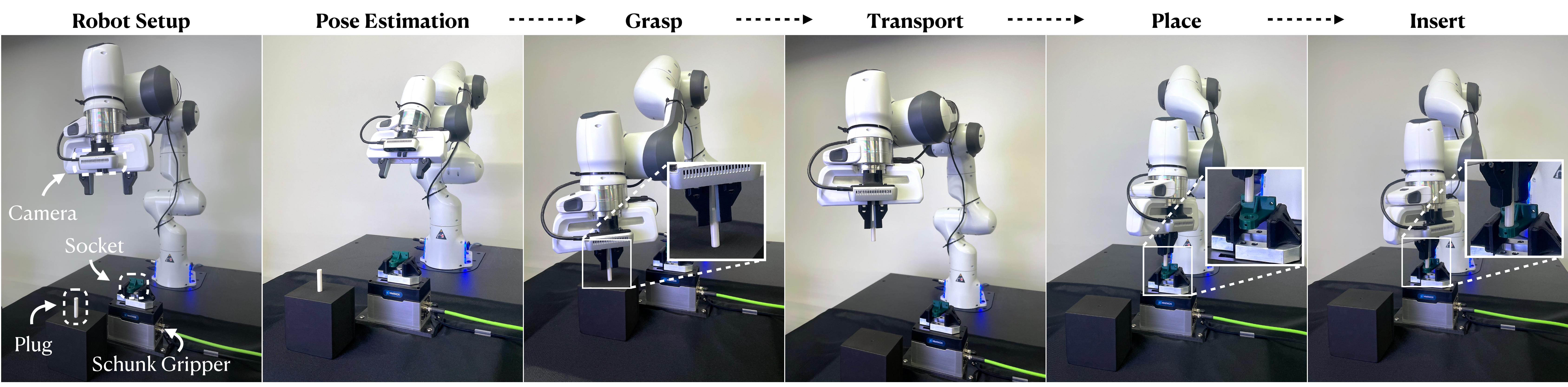}
    \caption{\textbf{Real-world perception-initialized assembly procedure.} We illustrate our procedure for performing perception-initialized assembly (i.e., from perception to insertion). Setup: We press a plug haphazardly onto a foam block, and we place a socket haphazardly within a Schunk gripper on a table. Pose Estimation: We estimate the pose of the plug or socket using our perception pipeline (\textbf{Section~\ref{sec:appendix_perception}}). Grasp: We grasp the plug using the output of our grasp optimizer (\textbf{Section~\ref{sec:appendix_grasp_optimization}}). Transport: We transport the plug across the workspace. Place: We place the plug above the socket. Insert: We deploy a specialist or generalist policy (\textbf{Section~\ref{sec:learning_methods}).}}
    \label{fig:task-sequence}
\end{figure*}

Our second evaluation question is, \textbf{do our generalist policies transfer to the real world?} For each of the same 20 assemblies, we deploy the generalist policy (trained from specialists for those assemblies) 10 times, for a total of 200 trials. \textbf{Figure~\ref{fig:eval_sim_real}} (right) shows our results.

The mean success rate in the real world is 84.50 $\pm$ 21.79\%, whereas the success rate in simulation is 80.42 $\pm$ 20.93\%.
For 16/20 assemblies, real-world success rates are higher.
Moreover, real-world success is within 2.0\% of that of the specialists.
Such results again indicate that our simulated training conditions are sufficiently adverse, and more importantly, that our distillation approach is highly effective.
Qualitatively, the generalist exhibits smoother motion than the specialists with identical gains, suggesting an averaging effect during training.

\subsection{Real-World Perception-Initialized Evaluation}
\label{subsec:real_world_end_to_end}

Finally, we present the results of our specialist and generalist policies as part of a perception-initialized assembly workflow.
For these trials, we place the plug haphazardly on a foam block and place the socket haphazardly within the Schunk gripper.
We capture an RGB-D image, estimate the poses of the parts, grasp the plug, transport it to the socket, and deploy a specialist or generalist  assembly policy (\textbf{Figure~\ref{fig:task-sequence}}).
This workflow presents a unique challenge, as initial part poses are even less constrained, and perception and/or control error accumulates at each stage, demanding increased policy robustness.

Our question is, \textbf{can our specialist and generalist policies help solve the perception-initialized assembly task?}
For 5 distinct assemblies from \textbf{Figure~\ref{fig:dataset_real}}, we deploy the corresponding specialists 10 times, for a total of 50 trials.
We repeat the procedure for the generalist policy.
\textbf{Table~\ref{tab:eval-ppi}} shows our results.

\begin{table}[!htb]
\rowcolors{3}{white}{Gainsboro}
\centering
\scalebox{0.95}{
\begin{tabular}{ l|cc|cc } 
\toprule
\multirow{2}{*}{Asset ID} & \multicolumn{2}{c}{\textbf{Specialist}} & \multicolumn{2}{c}{\textbf{Generalist}} \\ 
&  Policy-Only & Perception-Init & Policy-Only & Perception-Init \\ \midrule
00015  & 10/10 & 10/10 & 10/10 & 10/10\\ 
00296 & 8/10 & 8/10 & 10/10 & 9/10 \\ 
00320 & 10/10 & 10/10 & 10/10 & 8/10 \\ 
00340 & 10/10 & 9/10 & 10/10 & 8/10 \\ 
00346 & 9/10 & 8/10 & 8/10 & 8/10 \\ \midrule
Total \# & 47/50 & 45/50 & 48/50 & 43/50 \\ 
\textbf{Total} (\%) & \textbf{94.0\% }& \textbf{90.0\% }& \textbf{96.0\%} & \textbf{86.0\%}\\ 
\bottomrule
\end{tabular}}

\caption{\footnotesize{\textbf{Real-world evaluations for perception-initialized assembly.} We provide success rates of our specialist and generalist policies as part of a perception-initialized} assembly workflow. We also compare to isolated policy evaluations from \textbf{Section~\ref{subsec:real-world_results}}}.
\vspace{-2em}
\label{tab:eval-ppi}
\end{table}

For specialists, the mean success rate is 90.0\%, which is within 4.0\% of isolated policy evaluations from \textbf{Section~\ref{subsec:real-world_results}}, and for generalists, it is 86.0\%, which is within 10.0\%.
These results indicate that 6-DOF pose estimation, grasp optimization, and our proposed methods for learning specialist and generalist policies can be effectively combined to achieve perception-initialized assembly.
Qualitatively, higher success rates occur on assemblies with distinct visual features, as these correlate with more accurate pose estimates.
We conduct additional evaluations of the robustness of our policies to observation noise in \textbf{Appendix~\ref{sec:appendix_sim_to_real_robustness_obs_noise}}.

\section{Conclusions}

We present \textbf{AutoMate}, a learning framework and system for solving diverse assembly problems with specialist and generalist policies.
To our knowledge, AutoMate provides the first simulation-based framework for learning specialist and generalist policies over a wide range of assemblies, as well as the first system to demonstrate zero-shot sim-to-real over such a range.
We evaluate our framework and system over 5M+ simulated trials and 500 real-world trials.

A key limitation of our work is that for 20 assemblies, our specialist policies achieve $<$80\% success rates (\textbf{Figure~\ref{fig:results_specialist_all_assemblies}}). Policy failures are typically caused by unstable grasps on irregular geometries or sudden slip between the contact surfaces of the plug and socket. We anticipate that simple hardware improvements (e.g., higher-force grippers) and algorithmic improvements (e.g., utilization of demonstrations where the robot recovers from slip) may resolve these failure cases.

Our work opens up exciting opportunities for future work. First, we currently solve 2-part assemblies, which do not require sequence planning. In future work, we will develop an accelerated sequence planner for multi-part assemblies.

Second, our trajectory-matching approaches focus on paths in $\mathbb{R}^3$. We will generalize our trajectory-matching approaches (i.e., DTW and signature transforms) to paths consisting of $\text{SE}(3)$ transforms, facilitating assembly of parts requiring significant reorientation during alignment and insertion.

Third, we train specialist policies on a wide range of assemblies and distill a generalist policy from 20 specialists. We will continue to distill generalists from more specialists using powerful model architectures and larger model capacities, while consistently evaluating our results in the real world.

Through this line of work, we aim to gradually build towards a large-model paradigm for industrial robotics, while staying grounded in real-world deployment.

\bibliographystyle{plainnat}
\bibliography{references}

\pagebreak

\appendix
\renewcommand\thefigure{S\arabic{figure}}

\subsection{Problem Description: Real-World Experimental Setup}
\label{sec:appendix_real_world_experimental_setup}

\begin{figure}
    \centering
    \includegraphics[width=.35\textwidth]{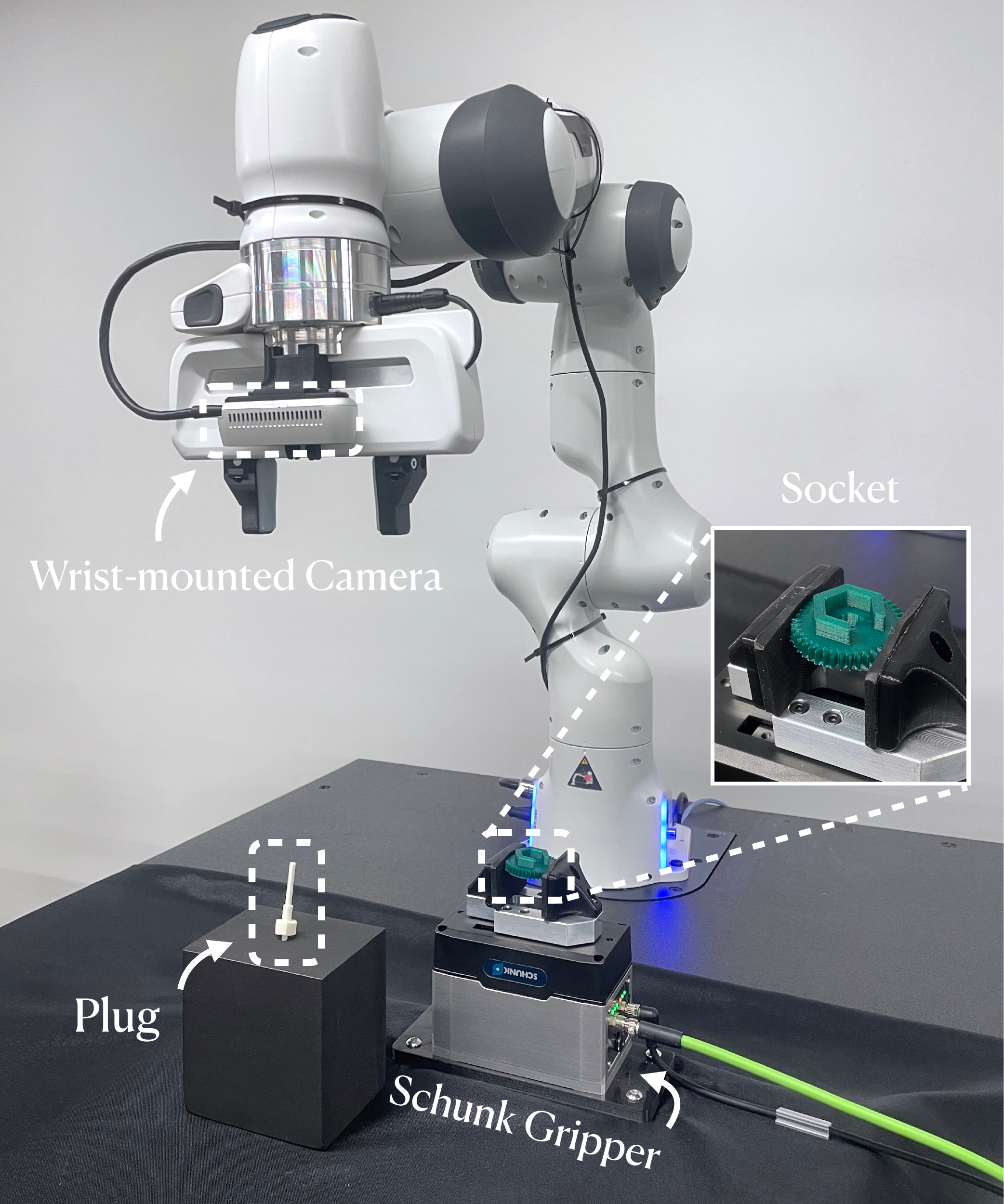}
    \caption{\textbf{Real-world experimental setup.} A Franka Panda robot (with a wrist-mounted Intel RealSense D435 camera) and a Schunk EGK40 gripper are mounted to a tabletop. At the beginning of each experiment, a 3D-printed plug is haphazardly pressed into a foam block or placed in the robot gripper, and a 3D-printed socket is haphazardly placed in the Schunk gripper. %
    }
    \label{fig:real-experimental-setup}
\end{figure}

\textbf{Figure~\ref{fig:real-experimental-setup}} shows our real-world experimental setup.

\subsection{Problem Description: Formal Problem Statement}
\label{sec:appendix_formal_problem_statement}

\noindent \textbf{Components:} %
\begin{itemize}
    \item A robotic manipulator $R$ with end effector $EE$
    \item A wrist-mounted RGB-D camera $C$
    \item A plug $P$ with known geometry described by mesh $M_P$
    \item A socket $S$ with known geometry described by mesh $M_S$
\end{itemize}

\noindent \textbf{Definitions:}
\begin{itemize}
    \item The configuration of robot $R$ is defined by joint angles $\Theta \in \mathbb{R}^7$.
    \item The pose of the end effector $EE$ is denoted by $X_{EE} \in SE(3)$.
    \item The robot $R$ is actuated by joint torques $T \in \mathbb{R}^7$.
    \item The camera $C$ captures an initial image pair $\{I^0, D^0\}$ consisting of an RGB image $I^0$ and a depth image $D^0$.
\end{itemize}

\noindent \textbf{Constraints:}
\begin{itemize}
    \item The relationship between the joint angles $\Theta$ and the end-effector pose $X_{EE}$ is governed by a forward-kinematics model $f$:
\[
X_{EE} = f(\Theta).
\]
    \item The relationship between the joint torques $T$, joint angles $\Theta$, and end-effector pose $X_{EE}$ is governed by a control law $g_\psi$:
\[
T = g_\psi\big(\Theta, X_{EE}, U(t)\big),
\]
where $\psi$ denotes constant parameters (e.g., control gains) and $U(t) \in \mathbb{R}^7 \cup SE(3)$ represents a control target in joint space ($\mathbb{R}^7$) or task space \big($SE(3)$\big) {at the current timestep $t$}.
{In our application, we use a task-space impedance controller} (\textbf{Equation~\ref{eq:task_space_impedance}}).
    \item {The initial poses of the end effector $EE$, plug $P$, and socket $S$ are given by}
    \begin{align*}
        X_{EE}^0 &\sim \text{Uniform}(a_{EE}^0, b_{EE}^0), \\
        X_P^0 &\sim \text{Uniform}(a_P^0, b_P^0), \\
        X_S^0 &\sim \text{Uniform}(a_S^0, b_S^0),
    \end{align*}
{where $a$ and $b$ denote user-defined bounds of the initial-pose distribution for the corresponding variable, and initial poses $X_P^0$ and $X_S^0$ satisfy the constraint that the convex hulls of meshes $M_P$ and $M_S$ do not overlap.
In our application, the bounds are specified in} \textbf{Table~\ref{tab:sim-randomization-ranges}}.

    \item {The goal pose for the plug $X_P^G$ satisfies the constraints that 1) the convex hulls of meshes $M_P$ and $M_S$ \textit{do} overlap, 2) meshes $M_P$ and $M_S$ do not intersect, and 3) user-specified surfaces on plug $P$ and socket $S$ are in contact (typically, an inferior surface of $P$ and superior surface of $S$).}
\end{itemize}

\noindent \textbf{{Goal:}} {The goal of the assembly task is to use joint angles $\Theta$, images $I^0$ and $D^0$, and meshes $M_P$ and $M_S$ to generate the control inputs $U(t)$ that guide plug $P$ from initial pose $X_P^0$ to goal pose $X_P^G$.}

\subsection{Methods: Mesh Preprocessing}
\label{sec:appendix_mesh_preprocessing}

We compile a dataset of assemblies based on Assemble Them All \cite{tian2022assemble}, which itself compiles a dataset based on the Fusion 360 Gallery dataset \cite{willis2022joinable}. The authors of \cite{tian2022assemble} preprocess the dataset from \cite{willis2022joinable} to ensure that the meshes are unique, normalized, watertight, and in a fully-assembled initial state, but the majority of the meshes still exhibit a minor degree of interpenetration in this state. Fortunately, the authors of \cite{tian2022assemble} use a simulator with penalty-based contact, which is robust to minor interpenetration \cite{xu2021end}; nevertheless, most widely-used robotics simulators (e.g., \cite{makoviychuk2021isaac}) enforce non-penetration constraints, which causes initially-interpenetrating parts to exhibit highly unstable dynamics. Furthermore, interpenetrating rigid parts cannot be assembled in the real world. Finally, the meshes from \cite{tian2022assemble} are unitless and have a wide range of relative sizes; for any particular choice of units, many are infeasible to manipulate with widely-used research robots. Thus, we preprocess meshes from \cite{tian2022assemble} such that they can be used in robotics simulators and assembled in reality.

Specifically, we preprocess each mesh as follows:

\begin{enumerate}
    \item \textbf{Scaling}: We choose units of meters, draw an oriented bounding box, and scale the mesh such that the longest edge of the bounding box is 10~cm, allowing it to be grasped by most robotic manipulators used in research.
    \item \textbf{Reorientation}: We reorient the mesh such that the primary axis of assembly (e.g., the insertion direction) is aligned with the global z-axis.
    \item \textbf{Translation}: We translate the mesh such that the bottom surface of the mesh is coplanar with the global origin when the mesh is in its assembled state.
    \item \textbf{Depenetration and Clearance}: If the mesh is a plug, we temporarily instantiate its corresponding socket. For each vertex on the plug, we compute its signed distance to the socket along the vertex normal using Warp \cite{macklin2022warp}; if the distance is negative (corresponding to interpenetration) or less than a desired radial clearance of 0.5 mm, we translate the vertex backward along its normal until achieving the desired clearance. Occasionally, this procedure produces unexpected results, such as when the plug is very thin or the socket is hollow; in such cases, we perform manual corrections in Blender.
    \item \textbf{Chamfering} (optional): We chamfer the contacting edges of the plug and socket using Blender. Chamfers are extremely common in assemblies, as they facilitate manual assembly, reduce stress concentrations, and remove burrs; for a cylindrical peg, chamfer sizes of $\frac{1}{10}$ to $\frac{1}{4}$ of the diameter are standard. As a typical plug in our dataset has a diameter $\sim$10~mm, we apply chamfers with a length of 1~mm and angle of 45 degrees. We provide chamfered and unchamfered versions of our meshes for use in simulation; in addition, we 3D print the chamfered versions for use in the real world, as our printer produces rough surfaces near curvature discontinuities.
    \item \textbf{Subdivision}: We subdivide the mesh using Trimesh \cite{dawson2019trimesh} until producing a minimum of 2000 vertices. During simulation, we use signed-distance-field (SDF)-based collisions \cite{macklin2020local, narang2022factory} as implemented in \cite{narang2022factory, makoviychuk2021isaac, nvidia2024isaac}; since this specific implementation generates a single contact per triangle, subdivision helps to ensure stable contact resolution along flat surfaces \cite{narang2022factorydocs}.
\end{enumerate}

\subsection{Methods: Grasp Optimization}
\label{sec:appendix_grasp_optimization}

The robot must first grasp the plug before assembling or disassembling the plug and socket, and the success of the assembly or disassembly process depends on the quality of the grasp. Thus, for each assembly, we perform an optimization procedure to determine a grasp that may lead to a high probability of success during assembly or disassembly. This {high-performing grasp} is then used when generating disassembly trajectories, training assembly policies, and deploying assembly policies in the real world for that assembly. Our grasp optimization procedure consists of the following steps:

\begin{figure}
    \centering
    \includegraphics[width=.45\textwidth]{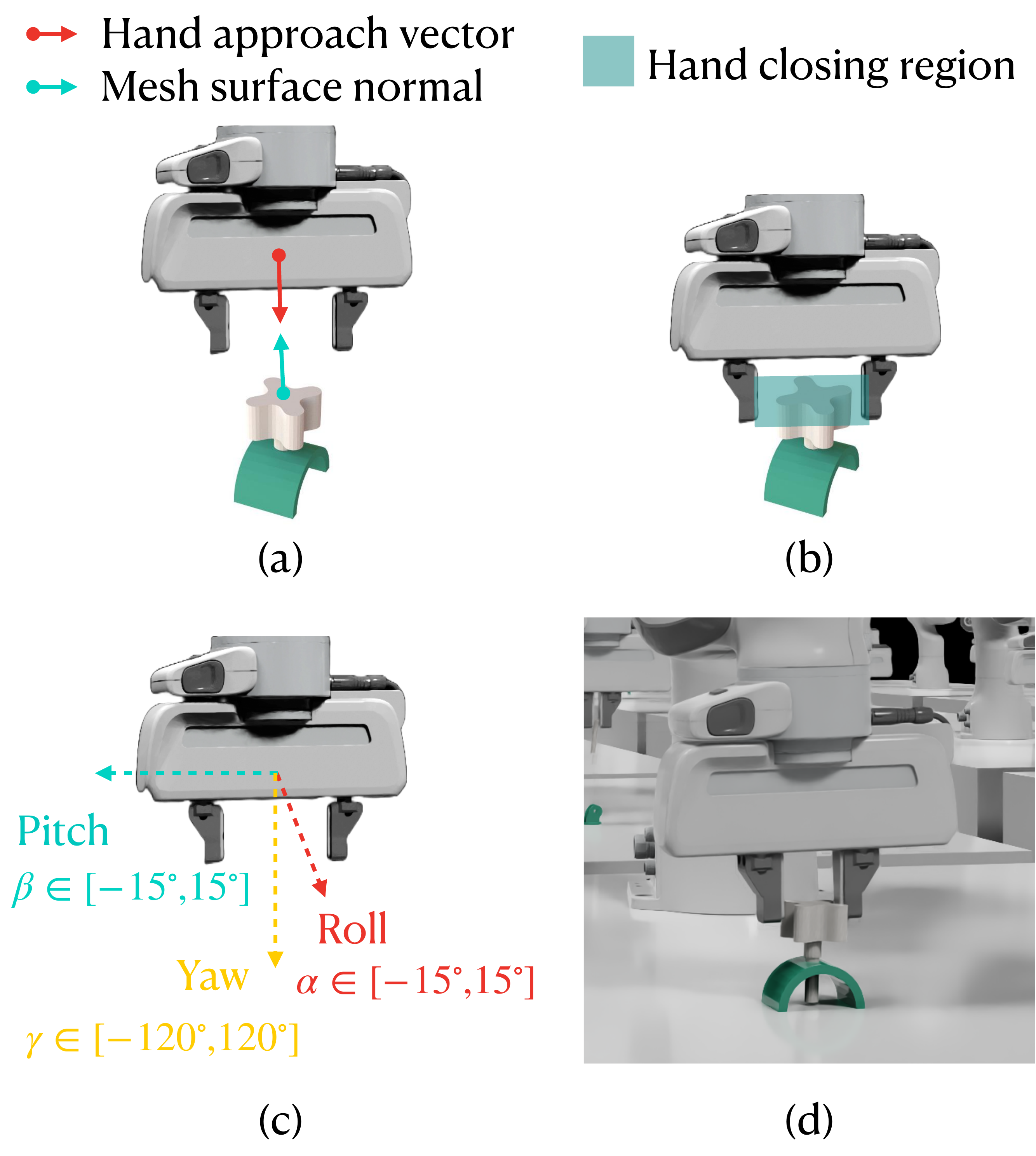}
    \caption{\textbf{Grasp sampling and evaluation procedure.} We apply a grasp sampling approach based on \cite{eppner2021acronym}, and we develop a a physics-based evaluation procedure. A) We sample a surface normal on the socket mesh, align the gripper axis with the normal, sample a position along the normal, and translate the gripper to that position. B) We check if the plug lies within the gripper closing region. C) We check if the Euler angles of the gripper are within specified bounds. D) We use simulation to disassemble the plug from the socket and check whether the plug remains in the gripper fingers.}
    \label{fig:grasp-sampling}
\end{figure}

\begin{enumerate}
    \item \textbf{Grasp Sampling} (\textbf{Figure~\ref{fig:grasp-sampling}}): We apply a kinematics- and geometry-based grasp sampling approach based on \cite{eppner2021acronym}. For each assembly, we first initialize the plug and socket meshes in their assembled state. We instantiate a robot gripper mesh, randomly sample a surface normal on the plug mesh, and align the central axis of the gripper to be collinear with the surface normal. We then randomly sample a position along this normal and translate the gripper to that position. We define a \textit{grasp sample} as the 6-DOF pose of the gripper in this state.

    We reject the grasp sample if 1) the robot hand intersects the plug or socket meshes, 2) the plug mesh does not intersect the gripper closing region (i.e., the prismatic volume contained between the fully-opened gripper fingers), or 3) the Euler angles of the gripper are outside of specified bounds ([-15, 15] deg for roll and pitch and [-120, 120] deg for yaw). The last of these criteria is designed to ensure that the Franka robot remains in a region of its workspace with high manipulability. We repeat this process until generating 100 grasp samples.
    
    \item \textbf{Physics-Based Evaluation}: Although the preceding grasp sampling procedure provides kinematically-feasible grasps, these samples are not guaranteed to be stable during the contact-rich interactions experienced during assembly and disassembly. Thus, we develop a subsequent physics-based evaluation phase.
    
    For each assembly, we first randomize the pose of the socket over a wide range (\textbf{Table~\ref{tab:sim-randomization-ranges}}), and we initialize the plug in its assembled state (i.e., inserted in the socket). For each of the 100 grasp samples, we execute the grasp on the plug. We use a task-space impedance controller \cite{lynch_modern_2017} to lift the plug from the socket until the convex hull of the plug no longer intersects the convex hull of the socket, and we move the robot gripper to a pose in free space randomly sampled from specified bounds ([-0.05, 0.05] for X- and Y-position, [0, 0.05] for Z-position, and [-10, 10] deg for roll, pitch, and yaw). We check whether the grasp is successful (i.e., if the plug remains in the gripper fingers until the end of the procedure). We repeat this procedure 1000 times. Finally, we identify the grasp sample with the highest success rate and designate that sample as the highest-performing grasp for the given assembly. In total, we run 10 million trials (100 assemblies x 100 grasps per assembly x 1000 trials per grasp), but we distribute the evaluations over many parallel environments for efficiency.
\end{enumerate}

Thus, the output of the grasp optimization procedure is a dictionary that maps each assembly to the highest-performing grasp for the corresponding plug. This grasp is inherently collision-free with respect to the socket in the assembled state, robust to large variations in robot configuration and plug/socket pose, and robust to contact-rich interactions.

\subsection{Methods: Disassembly Path Generation}

\begin{figure}
    \centering
    \includegraphics[width=.4\textwidth]{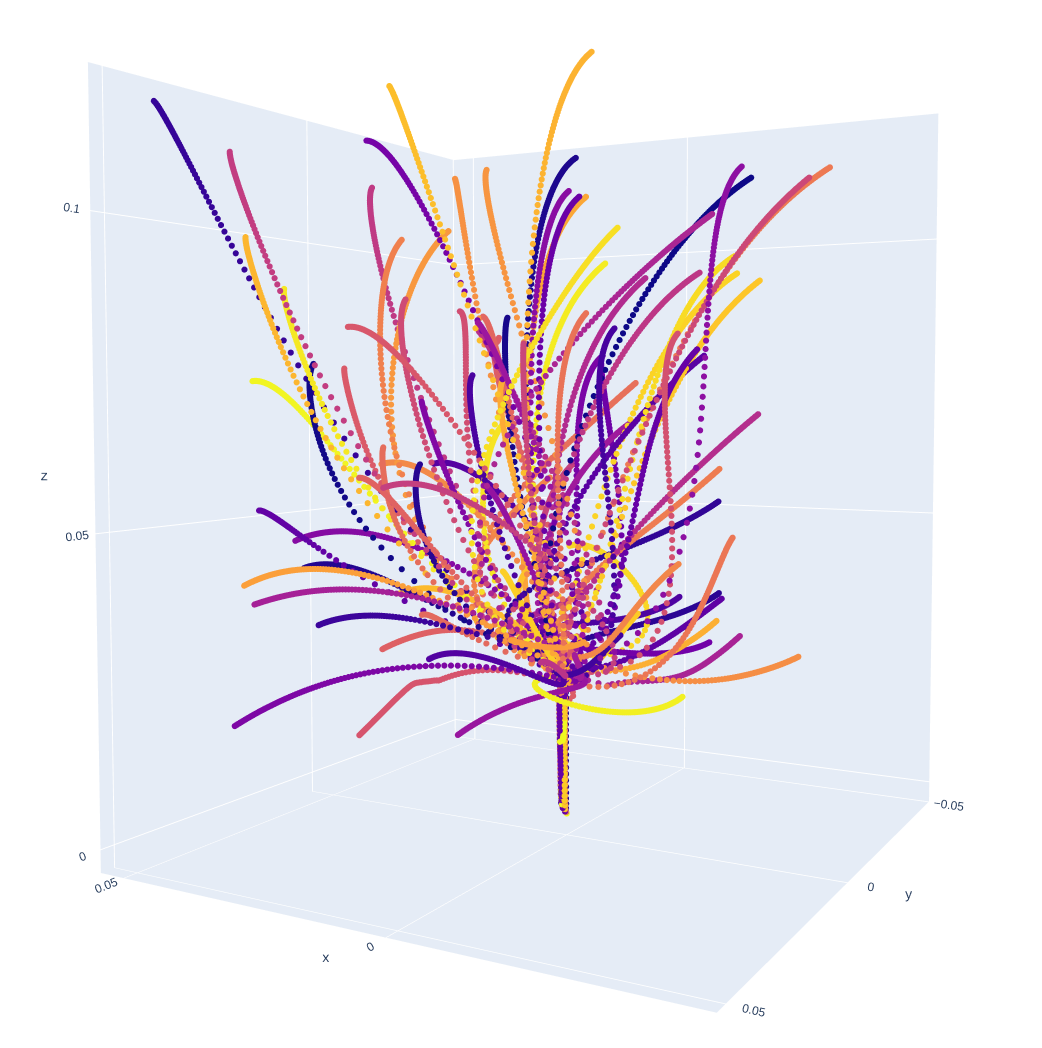}
    \caption{\textbf{Generated disassembly paths.} We generate disassembly paths via physics simulation and reverse the paths for use in assembly. Here we visualize 100 disassembly paths for an assembly with a deep socket.}
    \label{fig:disassembly-paths}
\end{figure}

\textbf{Figure~\ref{fig:disassembly-paths}} shows a visualization of disassembly paths for a representative assembly.

\subsection{Methods: Reinforcement Learning}
\label{sec:appendix_reinforcement_learning}

We formulate the robotic assembly problem as a Markov decision process (MDP), where the agent is a simulated robot, and the environment is a simulated environment containing the parts to be assembled. We define a state space $\mathcal{S}$, observation space $\mathcal{O}$, and action space $\mathcal{A}$. Our state-transition dynamics are defined by $\mathcal{T}: \mathcal{S} \times \mathcal{A} \rightarrow \mathcal{S}$, which is governed by the physical laws of rigid-body dynamics implemented in our simulator. We define a randomized initial state distribution $\rho_0$ (\textbf{Table~\ref{tab:sim-randomization-ranges}}) and reward function $R: \mathcal{S} \rightarrow \mathbb{R}$ with discount factor $\gamma \in (0, 1]$. We constrain our agents to execute actions over episodes of length $N$ timesteps, and we define shorter learning horizons of length $T$ timesteps. Finally, our return $G$ is defined as
\begin{equation}
    G(T) = \mathbb{E}_{\pi} \big[ \Sigma^{T-1}_{t=0}\gamma^t R(s_t) \big]
\end{equation}
In other words, the return is the expected sum of discounted rewards over the horizon. The objective
is to train a policy $\pi : \mathcal{O} \rightarrow \mathbb{P}(\mathcal{A})$ that maximizes the return.

\begin{table}[ht]
\centering
\begin{tabular}{ p{0.1cm}p{0.1cm}|p{3cm}|p{3cm} } 
\toprule
& & \textbf{Parameter} & \textbf{Randomization Range} \\\midrule
\multirow{6}{*}{\begin{turn}{90} Socket \end{turn}} & &  X-position (m) & [0.40, 0.60]  \\
& & Y-position (m) & [-0.10, 0.10]  \\
& & Z-position (m) & [0.16, 0.18] \\  %
& & roll angle (deg) & [-5, 5]  \\
& & pitch angle (deg) & [-5, 5] \\
& & yaw angle (deg) & [-5, 5] \\\midrule
\multirow{6}{*}{\begin{turn}{90} Plug  \end{turn}} & \multirow{6}{*}{\begin{turn}{90} (rel. to socket)  \end{turn}}  & X-position (mm)  & [-10, 10]  \\
&  & Y-position (mm) & [-10, 10] \\
&  & Z-position (mm) & [10, 20] \\  %
&  & roll angle (deg) & [-5, 5]  \\
&  & pitch angle (deg)& [-5, 5]  \\
&  & yaw angle (deg) & [-5, 5]  \\ \midrule
\multirow{6}{*}{\begin{turn}{90} Plug \end{turn}} & \multirow{6}{*}{\begin{turn}{90} (rel. to gripper) \end{turn}} & X-position (mm) & [-1, 1]  \\
& & Y-position (mm) & [-1, 1]  \\
& & Z-position (mm) & [-1, 1]  \\
& & roll angle  (deg) & [-5, 5]  \\
& & pitch angle  (deg) & [-5, 5]  \\
& & yaw angle  (deg) & [-5, 5]  \\ \bottomrule
\end{tabular}
\caption{\textbf{Randomization ranges for initial object poses during training.} We list position and orientation ranges for the following randomization procedure: First, the socket position and orientation are randomized. Next, the plug position and orientation are randomized relative to the socket pose. Then, the robot is commanded to move its gripper near the plug pose. Finally, the plug position and orientation are randomized again relative to the gripper pose. Values are all sampled from uniform distributions.}
\label{tab:sim-randomization-ranges}
\end{table}

To train policies, we use the proximal policy optimization (PPO) algorithm \cite{schulman2017proximal} due to its well-established performance over a wide range of simulation and sim-to-real problems \cite{makoviychuk2021isaac}, as well as its ease-of-use; we mitigate the low sample efficiency of PPO by using GPU-accelerated SDF-based contact simulation \cite{narang2022factory} and a GPU-accelerated PPO implementation \cite{rl-games2022}. We use PPO to learn a stochastic policy $\pi_{\theta}$ (i.e., an actor) parameterized by a neural network with weights $\theta$, as well as an approximation of the on-policy value function $V_{\phi} : \mathcal{S} \rightarrow \mathbb{R}$ (i.e., a critic) parameterized by a neural network with weights $\phi$. At evaluation and deployment time, the actor is deterministic, and the critic is neglected. For network architectures and hyperparameters, see \textbf{Table~\ref{tab:ppo-param}}.

\begin{table}[t]
\rowcolors{2}{white}{Gainsboro} 
\centering
\scalebox{0.9}{
\begin{tabular}{ l|c|c } 
\toprule
\textbf{Parameter} & \textbf{Specialist Training} & \textbf{Generalist Fine-tuning} \\ \midrule
MLP network size (actor) & [256, 128, 64] & [512, 256, 128, 64]\\
MLP network size (critic) & [256, 128, 64] & [256, 128, 64] \\
LSTM network size (actor) & 256 & 256 \\
Horizon length (T) & 32 & 32 \\
Adam learning rate & 1e-4 & 1e-4 \\
Discount factor ($\gamma$) & 0.99 & 0.99 \\
GAE parameter ($\lambda$) & 0.95 & 0.95 \\
Entropy coefficient & 0.0 & 0.0 \\
Critic coefficient & 2 & 2\\
Minibatch size & 8192 & 8192 \\
Minibatch epochs & 8 & 8 \\
Clipping parameter ($\epsilon$) & 0.2 & 0.2 \\ \bottomrule
\end{tabular}}
\caption{\textbf{Network architectures and hyperparameters used with Proximal Policy Optimization (PPO).} We list our specialist and generalist policy network architectures for the actor and critic networks, as well as our most critical PPO hyperparameters. We use PPO to train specialist policies from scratch, as well as fine-tune generalist policies after distillation as described in \textbf{Section~\ref{subsec:generalist_learning}}.}
\label{tab:ppo-param}
\end{table}

Our observation space provided to the actor consists of robot-arm joint angles [$\mathbb{R}^7$], the current pose of the end effector (i.e., the pose of the robot-gripper fingertips) [$\text{SE}(3)$], the goal pose of the end effector [$\text{SE}(3)$], and the pose of the end effector relative to the current pose [$\text{SE}(3)$] (\textbf{Table~\ref{tab:method-observation}}).
During training, the goal pose is simply the pose of the end effector when it is grasping the plug in its {highest-performing grasp pose} (\textbf{Appendix~\ref{sec:appendix_grasp_optimization}}) while the plug is inserted into the socket.
We avoid including joint velocities, end-effector velocities, and joint torques in the observation space, as these measurements exhibit substantial noise in the real world and can impede sim-to-real transfer; we also avoid including plug pose, as measuring this quantity typically requires tactile sensing \cite{bauza2023tac2pose}.

However, we adopt an asymmetric actor-critic strategy \cite{pinto2017asymmetric}, where the states provided to the critic include privileged information that is not provided to the actor, as the critic is only used for training and is not deployed in the real world.
Here, the states provided to the critic include joint velocities [$\mathbb{R}^7$], end-effector velocities [$\mathbb{R}^6$], and plug pose [$\text{SE}(3)$].
In addition, to capture real-world control error, perception error, and sensor noise, we apply uniformly-sampled noise to all observations of the position and orientation of the socket that are provided to the actor (\textbf{Table~\ref{tab:sim-noise}}), but do not apply noise to the corresponding states provided to the critic.

\begin{table*}[ht]
\centering
\rowcolors{1}{white}{Gainsboro} 
\begin{tabular}{ llcc } 
\toprule
\textbf{Input} & \textbf{Dimensions} & \textbf{Actor} & \textbf{Critic} \\ \midrule
Arm joint angles & 7 & \cmark & \cmark \\
Fingertip pose & 3 (position) + 4 (quaternion) & \cmark & \cmark \\
Target pose & 3 (position) + 4 (quaternion) & & \cmark \\
Target pose with noise & 3 (position) + 4 (quaternion) & \cmark &  \\
Relative target pose with noise & 3 (position) + 4 (quaternion) & \cmark &  \\
Arm joint velocities & 7 & & \cmark  \\
Fingertip linear velocity & 3 & & \cmark  \\
Fingertip angular velocity & 3 & & \cmark \\
Plug pose & 3 (position) + 4 (quaternion) &  & \cmark  \\
Relative target pose & 3 (position) + 4 (quaternion) & & \cmark \\
\bottomrule

\end{tabular}
\caption{\textbf{Inputs to the actor and critic for specialist policies.} We list observations provided to the actor, as well as observations and states provided to the critic. } 
\label{tab:method-observation}
\end{table*}

\begin{table}[ht]
\rowcolors{2}{white}{Gainsboro} 
\centering
\begin{tabular}{ l|l } 
\toprule
\textbf{Parameter} & \textbf{Noise Range} \\\midrule
Socket X-position & [-2, 2] mm \\ 
Socket Y-position & [-2, 2] mm \\
Socket Z-position & [-2, 2] mm \\
Socket roll angle & [-5, 5] deg \\ 
Socket pitch angle & [-5, 5] deg \\
Socket yaw angle & [-5, 5] deg \\ \bottomrule
\end{tabular}
\caption{\textbf{Noise ranges for observations of socket pose during training.} We list the ranges for position and orientation noise applied to observations of the socket pose, which is used to compute the goal pose of the end effector. Values were sampled from a uniform distributions.}
\label{tab:sim-noise}
\end{table}

Our action space consists of incremental pose targets [$\text{SE}(3)$], which represent the position and orientation difference between the current pose $x_c$ and the target pose $x_t$; we choose incremental targets rather than absolute targets in order to select from a small, bounded spatial range.
We pass these targets to a task-space impedance controller 
\begin{equation}
\label{eq:task_space_impedance}
    \tau = J^T \big( k_p (x_t \ominus x_c) - k_d \dot x_c \big)
\end{equation}
where $J \in \mathbb{R}^{6 \times 7}$ is the geometric Jacobian; $K_p \in \mathbb{R}^{6 \times 6}$ and $K_d \in \mathbb{R}^{6 \times 6}$ are diagonal matrices consisting of proportional and derivative gains, respectively; $\dot{x_c} \in \mathbb{R}^{6}$ is the velocity vector; and $x_t \ominus x_c$ computes the incremental pose target.
We use a task-space impedance controller to generalize actions across robot configurations and avoid using inertial matrices, which have not been precisely measured for our robot manipulator.
We superimpose a nullspace controller to softly constrain the robot to maintain a configuration with high manipulability, which can be compromised by elbow drift.

Finally, our baseline reward formulation (without imitation) is derived from \cite{tang2023industreal} and is composed of terms that penalize distance-to-goal, penalize simulation error, reward task difficulty, and reward success. Specifically, the reward
\begin{enumerate}
    \item penalizes distance-to-goal through an SDF-based reward, which computes the distance between the current plug pose and the goal (i.e., assembled) plug pose through SDF queries, which are less sensitive to object symmetries than keypoint-based distance queries,
    \item penalizes simulation error through a simulation-aware policy update (SAPU), which computes the maximum interpenetration distance at each timestep, weights the reward in inverse proportion to distance if it is less than a threshold, and does not update the reward otherwise,
    \item rewards task difficulty through a sampling-based curriculum (SBC), which increases the lower bound (but not the upper bound) of the range of initial-pose randomization as the agent becomes more proficient at the task, and weights the return in proportion to task difficulty.
\end{enumerate}
We defer precise descriptions to \cite{tang2023industreal} and implementation details to \cite{tang2023industrealsim}.
Whereas \cite{tang2023industreal} also rewarded success by providing a bonus at the end of every episode if a keypoint distance between the plug and its goal fell below a threshold on the final timestep, we instead reward success with a bonus at the end of every \textit{horizon} if the \textit{translational} distance between the plug and its goal falls below a threshold at \textit{any} timestep.

Precisely, our return over each horizon is given as
\begin{equation}
    G(T) = w_{SBC} \sum_{t=0}^{T-1} \big( \omega_{SAPU} (\omega_{SDF} R_{SDF} + \omega_I R_I) \big) + R_{succ}
\end{equation}
where $w_{SBC}$ is the weighting factor based on task difficulty, as determined by the SBC algorithm; $\omega_{SAPU}$ is the weighting factor based on simulation error, as determined by the SAPU algorithm; $R_{SDF}$ is the distance-to-goal reward, as determined by the SDF-based reward; $R_I$ is the imitation-based reward, as described in detail in the main text; $\omega_{SDF}$ and $\omega_I$ are hyperparameters to determine the relative importance of the distance-to-goal reward and the imitation-based reward; and $R_{succ}$ is a success bonus applied at the end of each horizon.
Parameters $\omega_{SDF}$ and $\omega_I$ are tuned simply so that $R_{SDF}$ and $R_I$ fall within the same order of magnitude.

\subsection{Methods: Dynamic Time Warping}
\label{sec:appendix_dynamic_time_warping}

At each timestep, we aim to determine the best reversed disassembly path for the robot to mimic.
More specifically, given the assembly path the robot has already traversed during the episode, we aim to select the closest disassembly path to imitate.
The first method we leverage for this procedure is dynamic time warping (DTW), which is described as follows:

Consider a time sequences $a = [a^1, a^2, ..., a^P]$, which might represent the path the robot has traversed, and a time sequence $b = [b^1, b^2, ..., b^Q]$, which might represent a disassembly path; we aim to compute the distance between these paths.
If we repeat this procedure for all disassembly paths, we can select the closest disassembly path as desired.

DTW matches each point $a_i$ to one or more points $b_j$, and vice versa. 
The matching process minimizes a cost $C(a, b)$, which is defined as the sum of a manually-defined distance function (typically, Euclidean distance) between each point $a_i$ and its match(es) from $b$. 
Moreover, the matching process satisfies the following constraints:

\begin{enumerate}
    \item Point $a_1$ must match with at least point $b_1$ (i.e., first points are aligned)
    \item Point $a_P$ must match with at least point $b_Q$ (i.e., last points are aligned)
    \item All matches must be monotonic (i.e., if point $a_i$ matches with point $b_j$, then point $a_{i+1}$ cannot match with point $b_{j-1}$ and point $a_{i-1}$ cannot match with point $b_{j+1}$)
\end{enumerate}

Ultimately, DTW returns the total distance between the optimal matches of sequence $a$ and sequence $b$.
We leverage the fast DTW implementation from \textit{Soft-DTW} \cite{cuturi2017soft}.

\textbf{Algorithm~\ref{alg:dtw}} provides pseudocode for a naive implementation of DTW. 
In this implementation, a matrix $M$ is constructed, where each $M[i][j]$ describes the minimum cost of matching $a[i]$ with $b[j]$. The implementation loops through each $M[i][j]$ and assigns its value to the distance between $a[i]$ and $b[j]$, plus the minimum accumulated cost of all previous possible matches. Importantly, only 3 such accumulated costs need to be considered: the accumulated costs of matching $a[i-1]$ and $b[j]$, $a[i]$ and $b[j-1]$, and $a[i-1]$ and $b[j-1]$. Intuitively, these are the only accumulated costs that 1) leave no previous point unmatched, and 2) are compliant with constraint 3. The value of element $M[P][Q]$ is the final value assigned in the loop and represents the minimum accumulated cost $C^*(a,b)$ over all possible matches between $a$ and $b$.

\begin{algorithm}
\label{alg:dtw}
\caption{Dynamic Time Warping (DTW)}
\begin{algorithmic}[1]
\Require Sequence $a$ of length $P$ and sequence $b$ of length $Q$
\Ensure DTW distance between $a$ and $b$
\Function{DTWDistance}{$a, b$}
    \State Define matrix $M$ of shape $(P+1, Q+1)$
    \State Initialize all elements of $M$ to $\infty$
    \State $M[0][0] \gets 0$
    \For{$i \gets 1$ to $P$}
        \For{$j \gets 1$ to $Q$}
            \State $d \gets \| a[i] - b[j] \|_2$
            \State $M[i][j] \gets d + 
            \min(M[i-1][j], M[i][j-1], M[i-1][j-1])$
        \EndFor
    \EndFor
    \State \Return $M[P][Q]$
\EndFunction
\end{algorithmic}
\end{algorithm}

\subsection{Methods: Signature Transform}
\label{sec:appendix_signature_transforms}

At each timestep, we aim to select the best reversed disassembly path for the robot to mimic. 
More specifically, given the assembly path the robot has already traversed during the episode, we aim to select the closest disassembly path to imitate.
The second method we explore for this procedure is the signature transform.
For an interactive introduction, see \cite{foster2022brief}, and for a detailed overview, see \cite{chevyrev2016primer}.

Consider a continuous-time $3$-dimensional path given by $X : [a, b] \rightarrow \mathbb{R}^3$.
For example, we can define the path $p(t)_{a,b} = (x(t), y(t), z(t))_{a,b}$, where $x(t)$, $y(t)$, and $z(t)$ might represent the $x$, $y$, and $z$ coordinates of the path the robot has already traversed for $t \in [a, b]$.
We can also consider a second path, which might represent a disassembly path; we aim to compute the distance between these paths by simply computing the L2 norm between their path signatures, which are defined next.
If we repeat this procedure for all disassembly paths, we can select the closest disassembly path as desired.

Focusing on the path $p(t)_{a,b}$, the path signature is given by the collection of all possible path integrals between $x(t)$, $y(t)$, and $z(t)$.
Specifically, the \textit{first level} of the path signature is
\begin{align}
    S_1(p(t))_{a,t} & = & \big( S_1(x(t))_{a,t}, S_1(y(t))_{a,t}, S_1(z(t))_{a,t} \big),
\end{align}
where
\begin{align}
    S_1(x(t))_{a,t} & = & \int_a^t dx(t) & = & x(t) - x(a) \\
    S_1(y(t))_{a,t} & = & \int_a^t dy(t) & = & y(t) - y(a) \\
    S_1(z(t))_{a,t} & = & \int_a^t dz(t) & = & z(t) - z(a)
\end{align}
In this case, there are 3 total path integrals, and each integral only involves a single coordinate of the path $p(t)$.

Next, the \textit{second level} of the path signature is
\begin{align}
    S_2(p(t))_{a,t} & = & \big( S_2(x(t), x(t))_{a,t}, S_2(x(t), y(t))_{a,t}, \nonumber \\
    & & ..., S_2(z(t), z(t))_{a,t} \big),
\end{align}
where
\begin{align}
    S_2(x(t), x(t))_{a,t} & = & \int_a^t S(x(t))_{a,t} dx(t) \\
    S_2(x(t), y(t))_{a,t} & = & \int_a^t S(x(t))_{a,t} dy(t) \label{eq:path_integral_interpretable} \\
    ... & = & ... \nonumber \\
    S_2(z(t), z(t))_{a,t} & = & \int_a^t S(z(t))_{a,t} dz(t)
\end{align}
In this case, there are 9 total path integrals, and each individual integral involves 2 coordinates of the path $p(t)$. (Note that when $a = 0$, \textbf{Equation~\ref{eq:path_integral_interpretable}} can be interpreted as the area under the curve when $x(t)$ is plotted against $y(t)$.) Further levels of the continuous-time path signature can be derived in similar fashion, where the $i$th level consists of $3^i$ path integrals.

The full continuous-time path signature $S(p(t))_{a,b}$ for $t \in [a, b]$ consists of the ordered set of all integrals. Specifically,
\begin{equation}
    S(p(t))_{a,b} = (1, S_1(p(t))_{a,b}, S_2(p(t))_{a,b}, ...)
\end{equation}
where the first element is equal to $1$ by convention. Conveniently, because path integrals are translation invariant (i.e., unaffected if the integrand is shifted by a constant) and reparameterization invariant (i.e., unaffected if the integrand traces its path slower or faster in time), path signatures inherit these properties. Thus, they are an extremely convenient representation for time-series data that may have translational offsets and/or disparate discretization or sampling schemes. Finally, the signature transform is simply the functional $T(p(t))_{a,b} : p(t)_{a,b} \rightarrow S((p(t))_{a,b}$ that takes a path as input and produces the path signature as output for $t \in [a,b]$.

As our data is not continuous, but discrete, we use the discrete-time form of the path signature. The first level $S_1(p[N])_{A,N}$, where $A$ and $N$ are the first and current timestep indices, respectively, can be expressed as
\begin{align}
    S_1(x[N])_{A,N} & = & \sum_{i=A}^{N-1} (x[i+1] - x[i]) & = & x[N] - x[A] \\
    S_1(y[N])_{A,N} & = & \sum_{i=A}^{N-1} (y[i+1] - y[i]) & = & y[N] - y[A] \\
    S_1(z[N])_{A,N} & = & \sum_{i=A}^{N-1} (z[i+1] - z[i]) & = & z[N] - z[A]
\end{align}
 The second level $S_2(p[N])_{A,N}$ can be expressed as
\begin{align}
    S_2(x[N], x[N])_{A,N} & = & \sum_{i=A}^{N-1} (x[i+1] - x[A])(x[i+1] - x[i]) \\
    S_2(x[N], y[N])_{A,N} & = & \sum_{i=A}^{N-1} (x[i+1] - x[A])(y[i+1] - y[i]) \\
    ... & = & ... \nonumber \\
    S_2(z[N], z[N])_{A,N} & = & \sum_{i=A}^{N-1} (z[i+1] - z[A])(z[i+1] - z[i])
\end{align}
As in continuous time, further levels of the discrete-time path signature can be derived in similar fashion, where the $i$th level consists of $3^i$ path summations.

Finally, the full discrete-time path signature $S(p[N])_{A,B}$ for $N \in [A, B]$, where $B$ is the last timestep index, consists of the ordered set of all summations. Specifically,
\begin{equation}
    S(p[N])_{A,B} = (1, S_1(p[N])_{A,B} , S_2(p[N])_{A,B}, ...)
\end{equation}

We leverage the fast, GPU-based signature transform implementation from Signatory \cite{kidger2021signatory}.

\subsection{Methods: Point-Cloud Autoencoder}
\label{sec:appendix_point_cloud_autoencoder}

Our specialist policies do not take part geometry as an observation, as geometry is constant for each policy and would not benefit policy learning.
However, our generalist policy \textit{does} take part geometry as an observation; unlike comparatively-imprecise tasks such as part reorientation \cite{chen2022system}, assembling a wide range of parts without knowledge of geometry would be exceedingly difficult.
At the same time, the meshes for our parts typically consist of 1000-5000 vertices and edges. We may consider A) taking mesh data directly as input to the policy or B) pretraining a network to extract a latent representation of mesh data and passing the latent vector to the policy.
We choose option B, as directly consuming mesh data would require an exceptionally-large observation space, and learning a latent representation and an assembly policy simultaneously would be computationally challenging.

Specifically, we train an autoencoder on a large set of meshes $M$. Each mesh $m_i \in M$ consists of $(V_i, E_i)$, where $V$ are the vertices and $E$ are the (undirected) edges.
At each iteration, we sample a batch of meshes $B \subset M$; for each $m_i \in B$, 
we sample a point cloud $P_i$ online, with each point $p_i^j \in P_i$ lying on the surface of $m_i$.
The point cloud $P_i$ is passed to a PointNet encoder \cite{qi2017pointnet} based on the implementation from \cite{mu2021maniskill} to produce a latent vector $z_i$.
Vector $z_i$ is passed to a fully-convolutional decoder based on the implementation from \cite{wan2023unidexgrasp++} to produce a reconstructed point cloud $Q_i$ (\textbf{Figure~\ref{sec:appendix_point_cloud_autoencoder}}).
The network is trained to minimize reconstruction loss, defined here as the chamfer distance between $P_i$ and $Q_i$:

\begin{align*}
   L_{\mathrm{CD}} &= \frac{1}{|P_i|}\sum_{p \in P_i}\min_{q \in Q_i} \left \| p-q \right \| ^2_2+\frac{1}{|Q_i|}\sum_{q \in Q_i}\min_{p \in P_i} \left \| p-q \right \|^2_2.
\end{align*}

In our final training procedure, $|M|$ = 1000 meshes from \cite{tian2022assemble},
$N$ = 2000, and $|z_i|$ =  32.

We briefly note two aspects of our training procedure that improve reconstruction accuracy: 1) We normalize the mean and variance of the vertices $V_i$ for each mesh $m_i$ prior to training, such that the network is not biased by a non-uniform distribution of mesh sizes, and 2) We increase the depth of the encoder relative to the decoder, such that the encoder can learn a more abstract latent representation, whereas the decoder is discouraged from overfitting to the input data.

\begin{figure}
\centering\includegraphics[width=.49\textwidth]{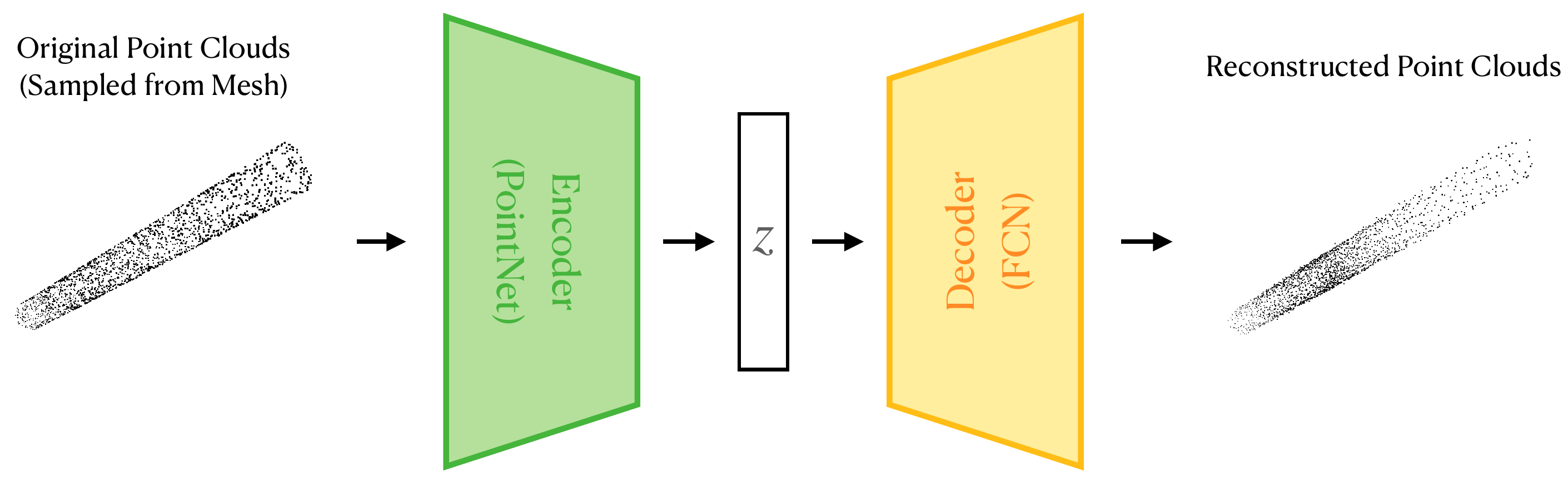}
    \caption{\textbf{Schematic of the point-cloud autoencoder.} We pass a point cloud as input to a PointNet encoder based on \cite{mu2021maniskill} to produce a latent vector $z$, which is in turn passed to a fully-convolutional decoder based on \cite{wan2023unidexgrasp++}. The autoencoder is trained to minimize reconstruction loss.}
    \label{fig:3d-autoencoder}
\end{figure}

Future work may focus on training an autoencoder with explicit or implicit surface information, which we hypothesize can improve the success rate of the generalist policy.
Possible methods include 1) augmenting each point $p_i^j$ with the local surface normal, 2) using a graph neural net (e.g., a graph convolutional network) that takes both points and edges as input \cite{ranjan2018generating, narang2021sim}, or 3) learning low-dimensional signed-distance-field (SDF) representations of the objects \cite{chou2022gensdf}.

\begin{figure*}
    \centering
    \includegraphics[width=0.9\textwidth]{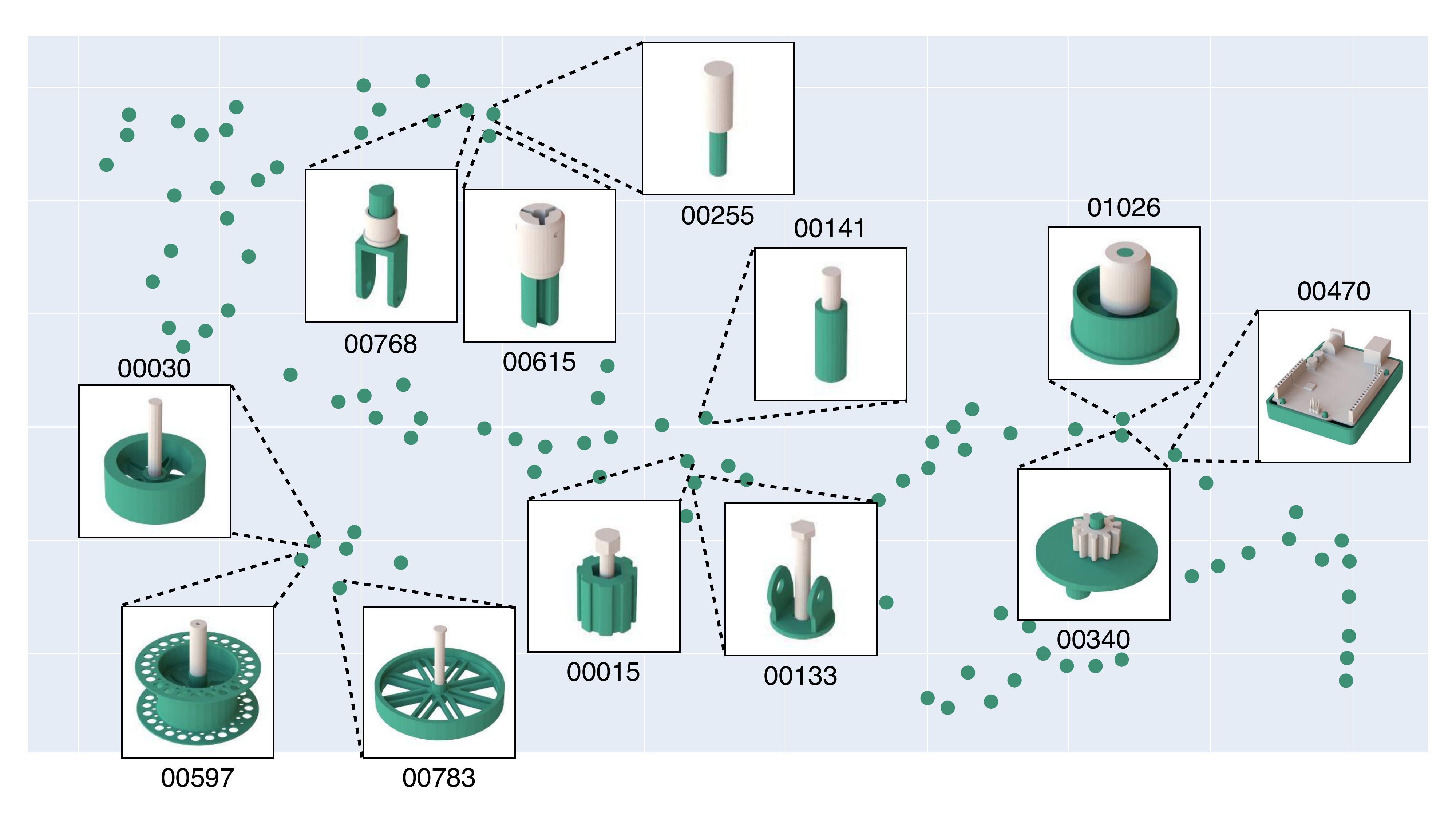}
    \caption{\textbf{t-SNE visualization of geometric representations of 100 assemblies.} As in \textbf{Figure~\ref{fig:t-sne_selected_assets}}, we plot the t-SNE representations of all 100 assemblies. Here, we show assemblies sampled from the same or nearby clusters; samples that are close in the lower-dimensional space have similar visual properties.}
    \label{fig:t-sne_clusters}
\end{figure*}

\subsection{Results: Specialist Policies (Continued)}
\label{sec:appendix_specialist_results}

\begin{figure*}
\centering\includegraphics[width=.98\textwidth]{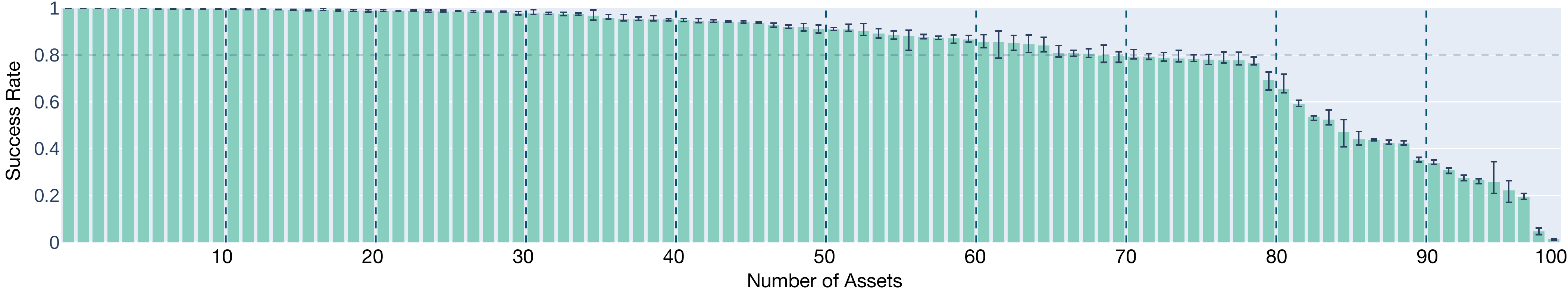}
    \caption{\textbf{Simulation-based evaluation of final training approach for specialist policies.} For each of the 100 assemblies, we train a specialist policy with the final AutoMate learning approach. For this approach, we train 5 random seeds, select the best seed, and evaluate it 5 times over 1000 trials. AutoMate maintains consistent performance across the majority of the assemblies and achieves the critical milestone of solving approximately 80\% of the assemblies with 80\% success rates or higher under substantial initial-pose randomization (\textbf{Table~\ref{tab:sim-randomization-ranges}}) and observation noise (\textbf{Table~\ref{tab:sim-noise}})}.
    \label{fig:results_specialist_all_assemblies}
\end{figure*}

\begin{figure*}
    \centering
    \includegraphics[width=\textwidth]{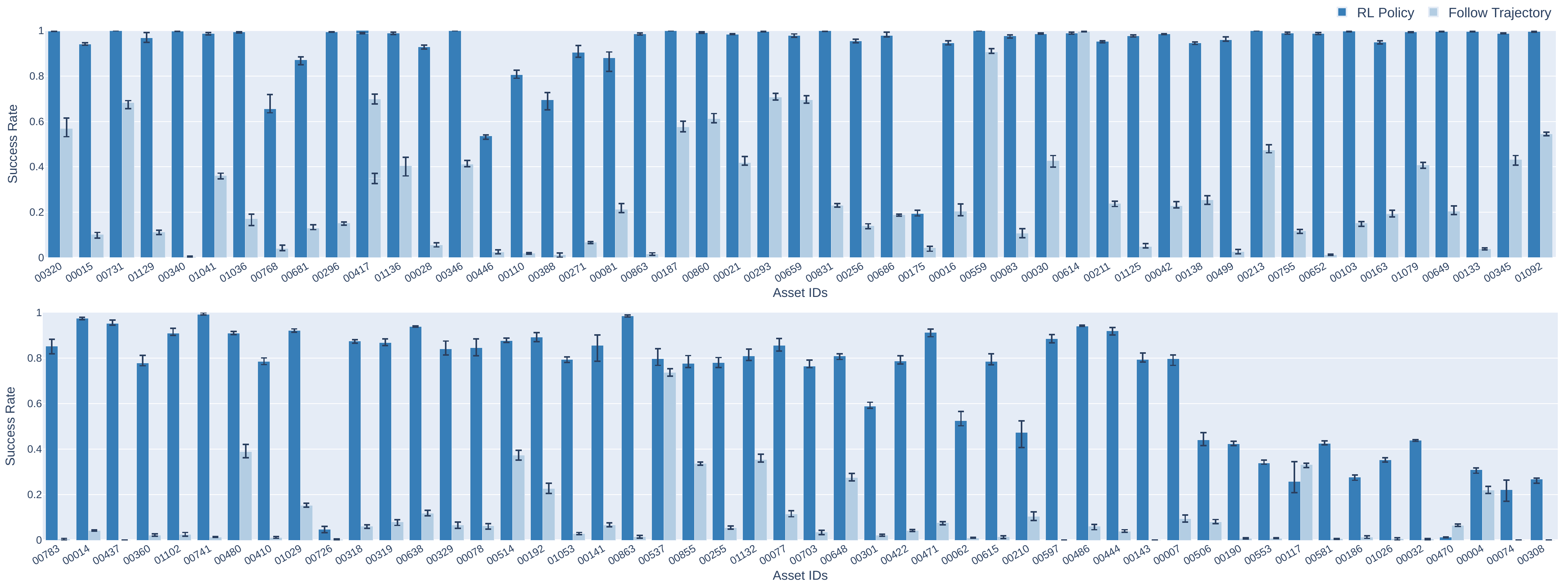}
    \caption{\textbf{Simulation-based comparison of final training approach for specialist policies with Follow Trajectory baseline.}
    {For each of the 100 assemblies, we evaluate both the final AutoMate learning approach and the Follow Trajectory method 5 times over 1000 trials.
    Each pair of bars shows a comparison between AutoMate and Follow Trajectory for a different assembly; the bottom plot is a continuation of the top plot.
    AutoMate significantly outperforms \textit{Follow Trajectory} on all assemblies except for assembly 00614, where the success rates are within 1\%.}}
    \label{fig:sim_rl_follow_traj_100}
\end{figure*}

\textbf{Figure~\ref{fig:results_specialist_all_assemblies}} shows the results of our final training approach for specialist policies over all 100 assemblies. \textbf{Figure~\ref{fig:sim_rl_follow_traj_100}} {shows an additional evaluation where the results are compared to the Follow Trajectory baseline} (\textbf{Section~\ref{subsec:specialist_results}}). {Our approach outperforms Follow Trajectory on 99 out of the 100 assemblies, typically by a significant margin.}

\subsection{Results: Robustness of Specialist Policies to Initial-Pose Randomization in Simulation}
\label{sec:appendix_specialist_robustness_initial_pose}

\begin{figure*}
    \centering
    \includegraphics[width=\textwidth]{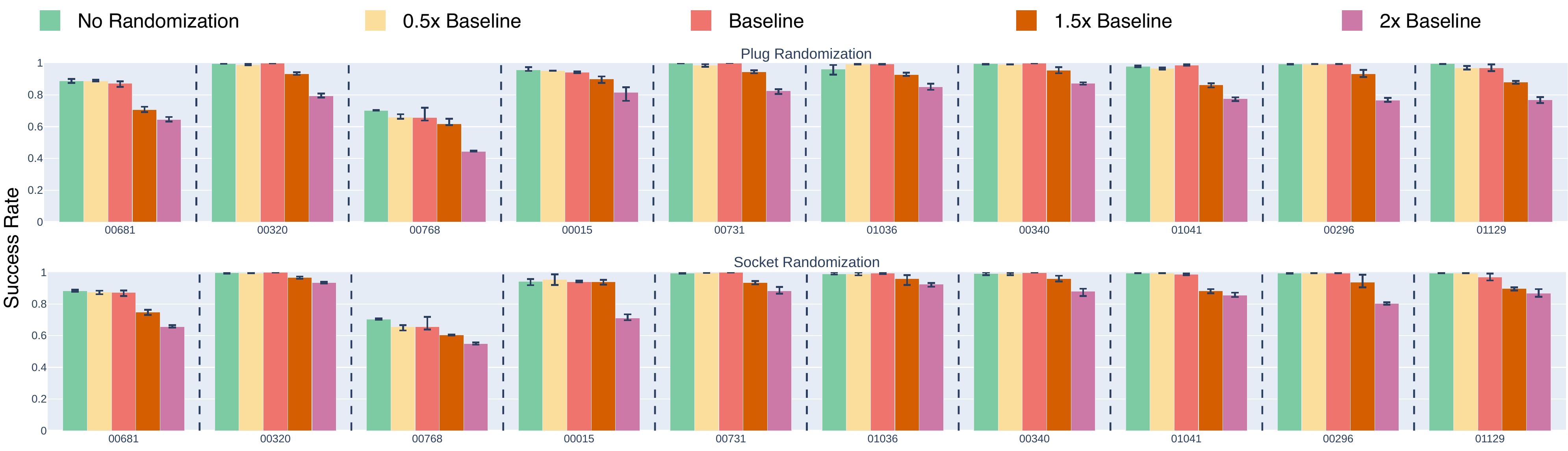}
    \caption{\textbf{Simulation-based evaluation of specialist policies with different levels of initial-pose randomization.}
    {For each assembly, we train our specialist policies with a baseline level of initial-pose randomization. We then evaluate our policies on in-distribution and out-of-distribution levels of randomization, indicated by the legend.
    For each experiment, we run 5 separate evaluations of 1000 trials each and compute mean success rates.} \textbf{Table~\ref{tab:sim_diff_rand_automate}} {provides the specific randomization ranges.}}
    \label{fig:sim_diff_rand_automate}
\end{figure*}

{We evaluate our specialist policies in simulation with different levels of initial plug- and socket-pose randomization for 10 assemblies.}
\textbf{Figure~\ref{fig:sim_diff_rand_automate}} {shows success rates at different randomization levels, and} \textbf{Table~\ref{tab:sim_diff_rand_automate}} {provides the corresponding randomization ranges and quantitative data.}
{The policies maintain high performance when tested in-distribution and moderately degrade when out-of-distribution (i.e., with larger initial-pose randomization than seen during training).}

\begin{table*}[ht]
\centering
\scalebox{0.75}{\begin{tabular}{ l|ccc|c|ccc|c } 
\toprule
Randomization & \multicolumn{4}{c}{\textbf{Socket Pose Randomization}} & \multicolumn{4}{c}{\textbf{Plug Pose Randomization (rel. to socket)}}   \\
 Range & XY Pos. (cm) & Z Pos. (cm) & Rot. (deg) & Success (\%) & XY Pos. (cm) & Z Pos. (cm) & Rot. (deg) & Success (\%) \\\midrule
\rowcolor{Gainsboro} No Randomization & 0 & 0 & 0 & 94.92$\pm$9.00 & 0 & 0 & 0 & 94.52$\pm$9.06 \\
0.5x Baseline & \([-5,5]\) & \([-0.5, 0.5]\) & \([-2.5,2.5]\) & 94.24$\pm$10.47 & \([-0.5, 0.5]\) & \([0,1]\) & \([-2.5,2.5]\) & 93.84$\pm$9.94\\
\rowcolor{Gainsboro} Baseline & \([-10,10]\) & \([-1.0,1.0]\) & \([-5.0,5.0]\) & 94.12$\pm$10.31 & \([-1.0,1.0]\) & \([0,2]\) & \([-5.0,5.0]\) & 94.12$\pm$10.31\\
1.5x Baseline & \([-15,15]\) & \([-1.5,1.5]\) & \([-7.5,7.5]\) & 88.21$\pm$11.34 & \([-1.5,1.5]\) & \([0,3]\) & \([-7.5,7.5]\) & 86.43$\pm$10.85\\
\rowcolor{Gainsboro} 2x Baseline &  \([-20,20]\) & \([-2.0,2.0]\) & \([-10.0,10.0]\) & 80.63$\pm$12.09 & \([-2.0,2.0]\) & \([0,4]\) & \([-10.0,10.0]\) & 75.56$\pm$12.01\\
\bottomrule
\end{tabular}}
\caption{\textbf{Simulation-based evaluation of specialist policies with different levels of initial-pose randomization.} {Here we provide randomization ranges and quantitative data corresponding to the plots in} \autoref{fig:sim_diff_rand_automate}. {Baseline refers to the level of initial-pose randomization under which the policies were trained. The left column lists all levels of randomization under which the policies were tested.}}
\label{tab:sim_diff_rand_automate}
\end{table*}

\begin{figure*}
    \centering
    \includegraphics[width=\textwidth]{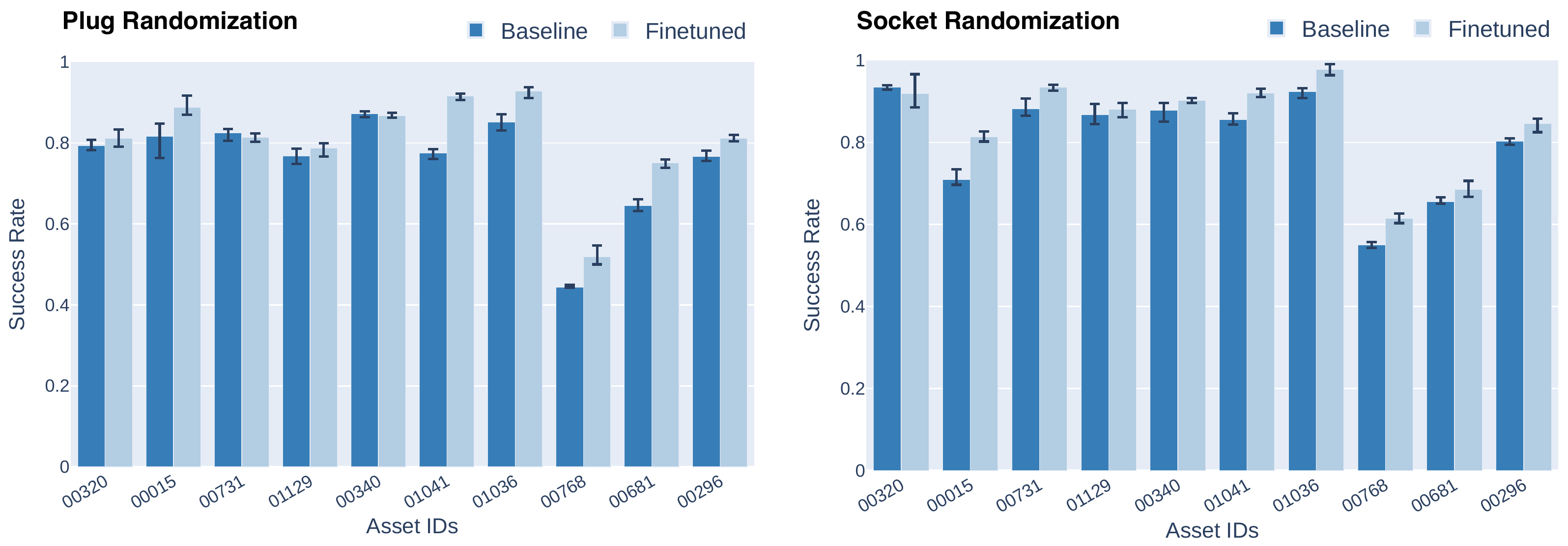}
    \caption{\textbf{Simulation-based evaluation of specialist policies fine-tuned with increased initial-pose randomization.} {For each assembly, we fine-tune the corresponding specialist policy with twice the initial plug- and socket-pose randomization seen during training; we then evaluate our policies over this larger range. For each policy, we run 5 separate evaluations of 1000 trials each and compute mean success rates. We compare to the non-fine-tuned policy.}}
    \label{fig:sim_finetune_rand}
\end{figure*}

{Nevertheless, out-of-distribution generalization is a challenge for nearly all learning-based methods; one simple and effective remedy is to simply fine-tune the policies on out-of-distribution data.
Thus, in simulation, we fine-tune our policies with twice the initial plug- and socket-pose randomization seen during original training, and we evaluate the policies over this larger range.}
\textbf{Figure~\ref{fig:sim_finetune_rand}} {shows the resulting success rates.
As predicted, fine-tuning is an effective strategy for allowing our policies to adapt to larger initial-pose randomization.} 

\subsection{Results: Robustness of Specialist Policies to Observation Noise in Simulation}
\label{sec:appendix_specialist_robustness_obs_noise}

\begin{figure*}
    \centering
    \includegraphics[width=\textwidth]{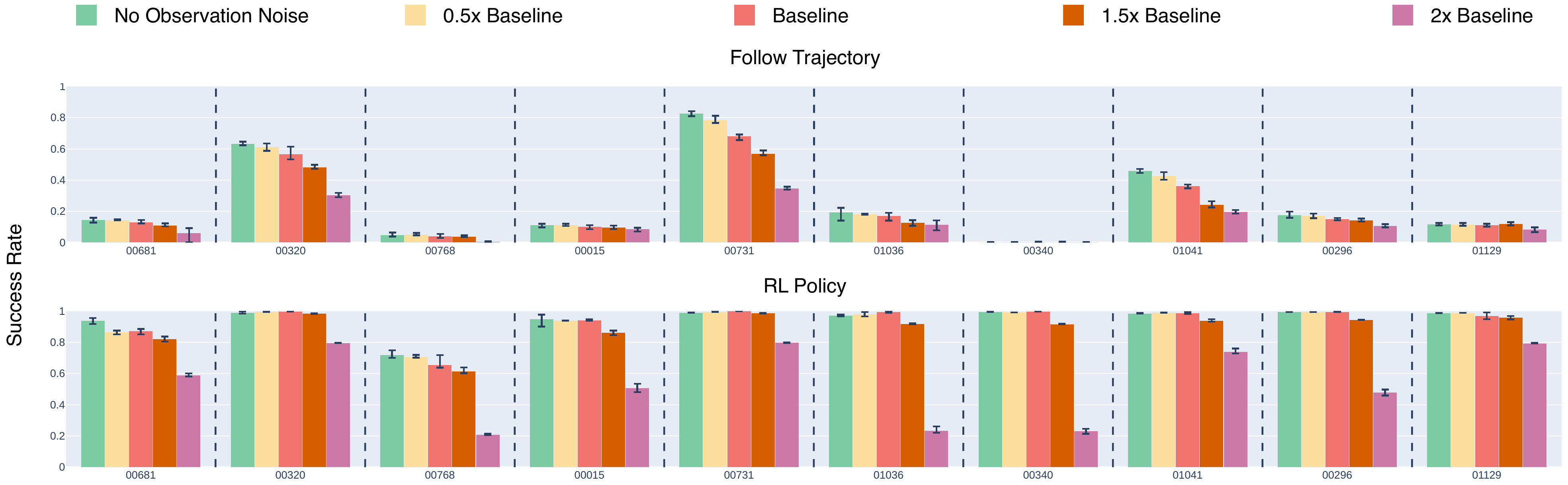}
    \caption{\textbf{Simulation-based evaluation of specialist policies with different levels of observation noise.} {(Bottom) For each assembly, we train our specialist policies with a baseline level of observation noise. We then evaluate our policies on in-distribution and out-of-distribution levels of observation noise, indicated by the legend.} \textbf{Table~\ref{tab:sim_diff_obs_10}} {provides the specific noise ranges. (Top) For instructive purposes, we also evaluate the Follow Trajectory approach under the same conditions. For all experiments, we run 5 separate evaluations of 1000 trials each and compute mean success rates.}}
    \label{fig:sim_diff_obs_10}
\end{figure*}

{We evaluate our specialist policies with different levels of observation noise applied to the socket pose for 10 assemblies.
For instructive purposes, we also compare the results to the Follow Trajectory baseline} (\textbf{Section~\ref{subsec:specialist_results}}).
\textbf{Figure~\ref{fig:sim_diff_obs_10}} {illustrates our results, and} \textbf{Table~\ref{tab:sim_diff_obs_10}} {provides the corresponding noise levels and quantitative data. The policies significantly outperform the Follow Trajectory baseline over all assemblies. In addition, the policies maintain high performance when tested in-distribution and moderately degrade when out-of-distribution (i.e., with larger observation noise than seen during training), particularly as the magnitudes of observation noise approach critical length scales of the plug and socket.}

\begin{figure}
    \centering
    \includegraphics[width=0.45\textwidth]{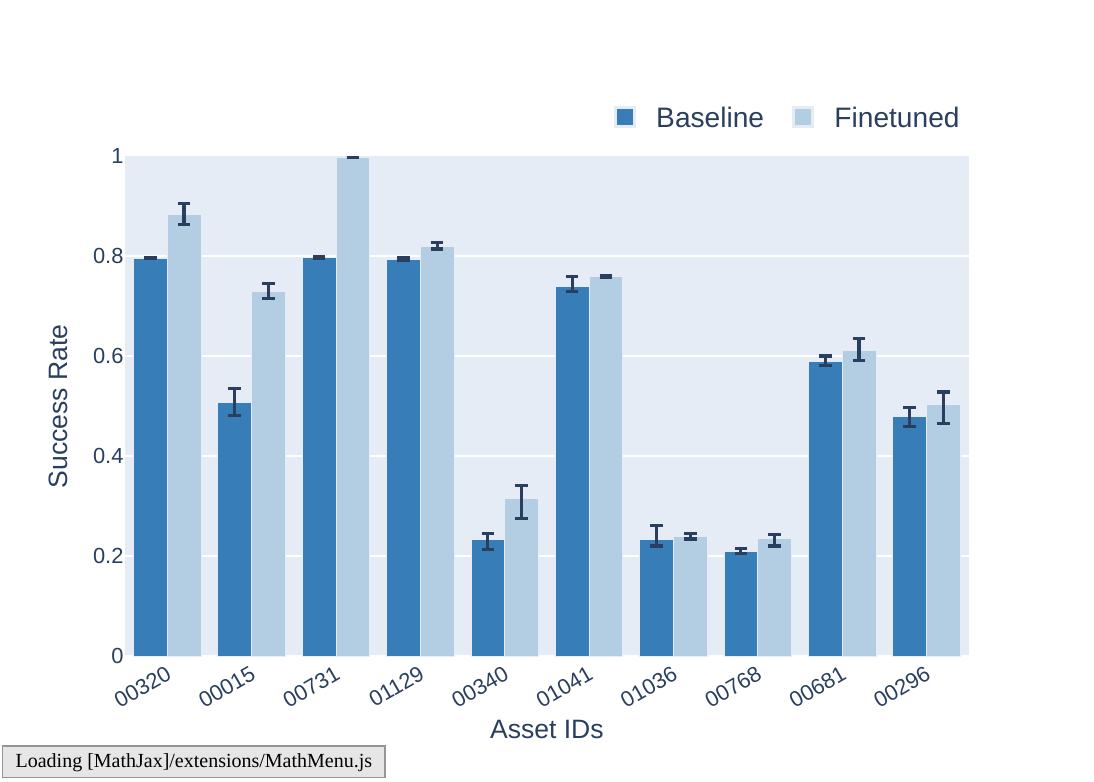}
    \caption{\textbf{Simulation-based evaluation of our specialist policies fine-tuned with higher level of observation noise.} {For each assembly, we fine-tune the corresponding specialist policy with twice the amount of observation noise seen during training; we then evaluate the policies over this larger range. For each policy, we run 5 separate evaluations of 1000 trials each and compute mean success rates. We compare to the non-fine-tuned policy.}}
    \label{fig:finetuned_obs_noise}
\end{figure}

{As with initial-pose randomization, we explore whether a fine-tuning strategy can improve performance. Specifically, in simulation, we fine-tune our policies with twice the observation noise seen during original training, and we evaluate the policies over this larger range.} \textbf{Figure~\ref{fig:finetuned_obs_noise}} {shows the resulting success rates. As demonstrated, fine-tuning can moderately improve performance under higher levels of observation noise.}

\begin{table*}[ht]
\centering
\scalebox{0.8}{\begin{tabular}{ l|cc|c|c } 
\toprule
Obs. Noise & \multicolumn{2}{c}{Observation Noise} & \multirow{2}{*}{RL Policy Success (\%)} & \multirow{2}{*}{Follow Trajectory Success (\%)}   \\
Level & Pos. (mm) & Rot. (deg) & & \\\midrule
\rowcolor{Gainsboro} No noise & 0 & 0 & 95.72$\pm$7.39 & 27.20$\pm$26.04\\
0.5x Baseline & \([-1,1]\) & \([-2.5,2.5]\) & 94.36$\pm$8.89 & 26.05$\pm$24.66 \\
\rowcolor{Gainsboro}Baseline &\([-2,2]\) & \([-5.0,5.0]\) & 94.12$\pm$10.31 & 23.25$\pm$21.83\\
1.5x Baseline &\([-3,3]\) & \([-7.5,7.5]\) & 90.63$\pm$9.07 & 19.42$\pm$17.84\\
\rowcolor{Gainsboro}2x Baseline &\([-4,4]\) & \([-10.0,10.0]\) & 51.24$\pm$26.61 & 13.14$\pm$11.20\\
\bottomrule
\end{tabular}
}
\caption{\textbf{Simulation-based evaluation of specialist policies with different levels of observation noise.} {Here we provide noise ranges and quantitative data corresponding to the plots in} \textbf{Figure~\ref{fig:sim_diff_obs_10}}. {Baseline refers to the level of noise under which the policies were trained. The left column lists all levels of noise under which the policies were tested. We also provide a comparison with the Follow Trajectory approach.}}
    \label{tab:sim_diff_obs_10}
\end{table*}

\subsection{Results: Generalist Policies}
\label{sec:appendix_generalist_results}

As a supplementary evaluation question, we ask, \textbf{what is the scaling law between generalist performance and the number of specialists used in training?}

\begin{figure}
    \centering
    \includegraphics[width=.48\textwidth]{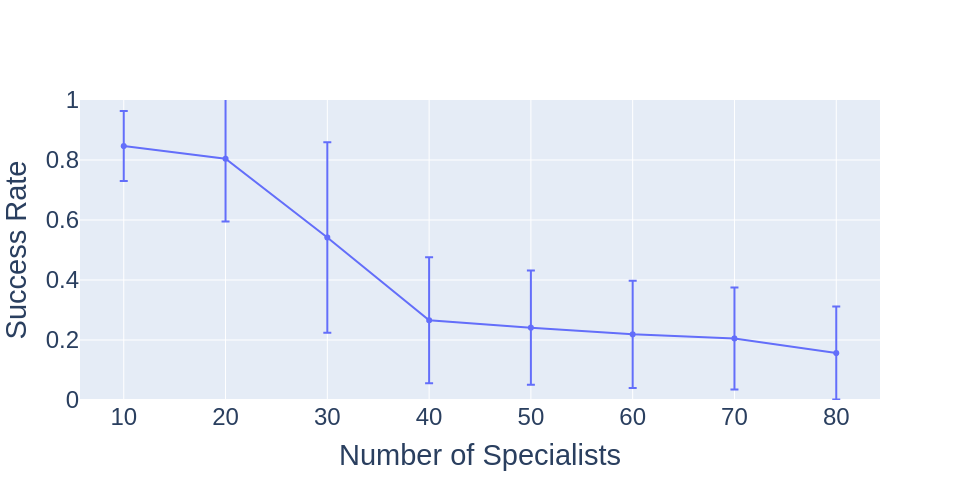}
    \caption{\textbf{Simulation-based evaluation of scaling law for training generalist policies.} We consider batches of \{10, 20, ..., 80\} assemblies, where each batch is evenly sampled in t-SNE space (\textbf{Figure~\ref{fig:t-sne_clusters}}). For each batch, we train a generalist policy with the final AutoMate learning approach, from all specialists corresponding to that batch of assemblies. We evaluate each generalist policy over all the assemblies in its corresponding batch over 5000 trials. We observe high success rates for a generalist trained over 10 and 20 assemblies ($\approx$80\%), a steep drop at 30 and 40 assemblies ($\approx$55\% and $\approx$30\%), and consistent, low success rates for more assemblies ($\approx$20\%).}
    \label{fig:scaling_law}
\end{figure}

To answer this question, we consider batches of \{10, 20, ..., 80\} assemblies, where each batch is evenly sampled in t-SNE space (\textbf{Figure~\ref{fig:t-sne_clusters}}).
For each batch, we train a generalist policy with \textit{RL + DAgger + RL (w/SBC)}, from all specialists corresponding to that batch of assemblies.
We evaluate each generalist policy over all the assemblies in its corresponding batch over 5000 trials, for a total of 1.8M trials.
\textbf{Figure~\ref{fig:scaling_law}} shows our results. We observe high success rates for a generalist trained from 10 and 20 specialists ($\approx$80\%), a steep drop for a generalist trained on 30 or 40 specialists ($\approx$55\% and $\approx$30\%), and consistent, low success rates for a generalist trained on more specialists ($\approx$20\%).
In future work, we aspire to formulate methods that can preserve generalist performance as the number of assemblies increases.

\subsection{Methods: Perception}
\label{sec:appendix_perception}

\subsubsection{Camera Calibration and Tuning} We observe the environment with a single wrist-mounted Intel RealSense D435 RGB-D camera.
It is a common sentiment among robotics researchers that off-the-shelf cameras may not be sufficient for high-precision tasks; moreover, it is common practice among the robot learning community to compensate for the weaknesses of such cameras primarily via data-driven strategies (e.g., increased data collection, data augmentation, etc.).
However, we have found that diligent camera calibration and tuning can greatly improve the RGB image quality, point cloud quality, and performance in downstream perception modules (e.g., pose estimators), to a level sufficient for high-precision assembly of parts far smaller than the robot manipulator.

Specifically, we take the following steps:

\begin{itemize}
    \item \textbf{Extrinsics calibration}: We calibrate our \textit{absolute} extrinsics (i.e., the pose of our RGB camera in the robot frame) using the procedure described in \cite{tang2023industreal}, with a lightly-modified implementation of the corresponding code in \cite{tang2023industreallib}. The procedure consists of moving the end effector to randomized target poses, capturing an image of an AprilTag in each pose, computing the pose of the AprilTag in the camera frame from each image, and using the Tsai-Lenz algorithm \cite{tsai1989new} to compute the extrinsics matrix. We do not calibrate our \textit{relative} intrinsics (i.e., the pose of our depth camera with respect to our RGB camera) and rely on the RealSense-provided matrix.
    \item \textbf{Intrinsics calibration}: We calibrate our intrinsics matrix using the procedure described in the RealSense whitepaper for on-chip self-calibration \cite{realsense2023selfcalibration}. The procedure consists of capturing an image of a textured target and running the manufacturer-provided calibration function.
    \item \textbf{Camera settings}: We tune our camera settings using the suggestions provided in the RealSense whitepapers on tuning depth cameras \cite{realsense2023tuningdepth} and depth image post-processing \cite{realsense2023depthpost}. The most impactful settings were
    \begin{itemize}
        \item \underline{RGB and depth camera exposure}: We tune the exposure to maximize the quality of the color and depth map images on our assemblies, rather than using the default autoexposure. Furthermore, we tune the RGB and depth camera exposures simultaneously (i.e., set them to the same value), rather than separately.
        \item \underline{Laser power}: We increase power from its default value to increase the density of the depth image.
        \item \underline{Spatial hole-filling}: We apply a hole-filling filter in post-processing to repair holes in the depth image.
    \end{itemize}
\end{itemize}

Finally, to optimize the performance of our downstream pose estimator (described next), we maximized RGB camera resolution (in order to increase the number of pixels on small surfaces), captured images from angled (rather than overhead) view, and avoided direct lighting of the assemblies.

\subsubsection{Pose Estimation}

In simulation, we train RL policies from 6D poses of the parts rather than from RGB images or point clouds, which would substantially increase compute requirements. On the other hand, in the real world, we observe the environment using a single Intel RealSense D435 RGB-D camera mounted on the wrist of the robot. Thus, in order to deploy our simulation-trained policies in the real world, we may consider A) moving simulation towards reality (i.e., distilling the simulation-trained policies to use RGB-image and/or point-cloud inputs) or B) moving reality towards simulation (i.e., extracting 6D poses from real-world RGB images and/or point clouds). We choose option B for several reasons: 1) We can avoid a cross-modal distillation process, 2) Simulated and real-world RGB images have a substantial sim-to-real gap, 3) The quality of real-world point clouds is poor on our small, mildly-reflective parts, and 4) The accuracy of pose estimators has improved dramatically in recent years \cite{labbe2020cosypose, labbe2022megapose, wen2023foundationpose}.

We assume that each part of each assembly has a known CAD model (specifically, an OBJ file with no associated texture), which is typical in industrial assembly settings. We use a pose estimation pipeline that takes as input 1) an RGB-D image of a part in the real world, 2) the camera intrinsics matrix, 3) the camera extrinsics matrix, and 4) a CAD model of the part, and then predicts the 6D pose of the part. This pose can subsequently be passed as input to our RL policies.

Our pose estimation pipeline consists of the following steps:
\begin{enumerate}
    \item \textbf{Image capture}: The RealSense camera is used to capture a 1280 x 720 RGB image and 1280 x 720 depth image of a part with a known CAD model.
    \item \textbf{Part selection}: The RGB image is shown to the user. The user can left-click on the part to provide a positive annotation (i.e., a pixel that lies on the part of interest).
    \item \textbf{Segmentation}: The RGB image, pixel location(s), and annotation(s) are passed to \cite{kirillov2023segment}, which produces a high-accuracy segmentation mask for the part.
    \item \textbf{Refinement (optional)}: If the mask does not span the part, the user can provide another positive annotation; conversely, if the mask includes background features, the user can right-click to provide a negative annotation (i.e., a pixel that does not correspond to the part). Segmentation is then executed with the additional annotations.
    \item \textbf{Model-based estimation}: The RGB image, depth image, camera intrinsics, segmentation mask, and textureless CAD model are fed to \cite{wen2023foundationpose}, which regresses to the 6D object pose in the camera frame.
    \item \textbf{Frame transformation}: The 6D object pose in the camera frame is combined with the camera extrinsics to compute the 6D object pose in the robot frame.
\end{enumerate}
In total, the steps above take $\approx$20 seconds to execute.

Future work may focus on replacing the click-based interface for selecting parts with a language interface, using faster implementations of the segmentation model (e.g., \cite{zhao2023fast, zhang2023faster}), automatically retrieving the appropriate CAD model from a database, and optimizing \cite{wen2023foundationpose} for faster performance.

\begin{figure*}
    \centering
    \includegraphics[width=\textwidth]{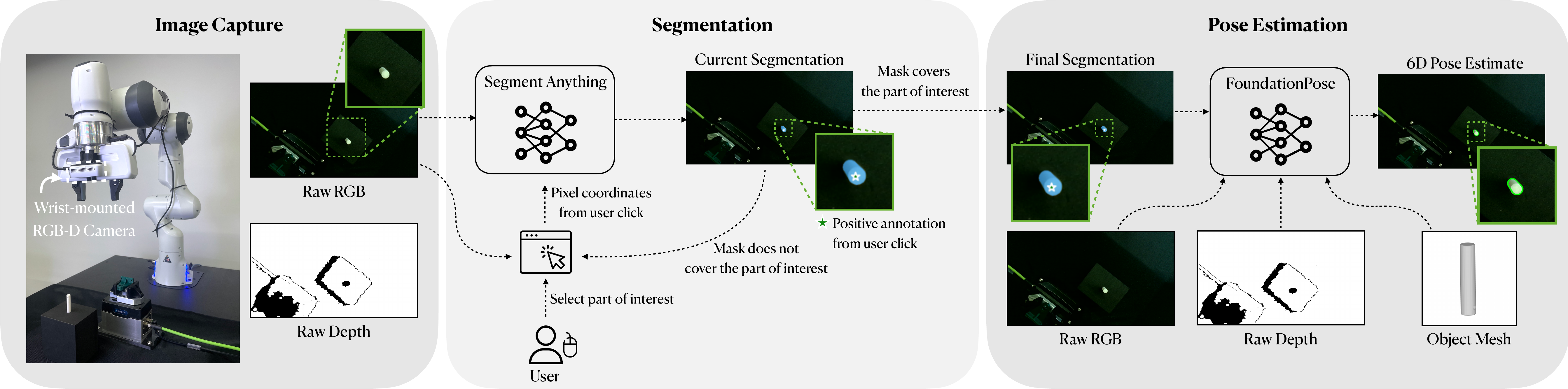}
    \caption{\textbf{Real-world perception pipeline.} Left: At the beginning of our pipeline, we use an Intel RealSense D435 RGB-D camera mounted on the wrist of the robot to capture an RGB image and depth image. Middle: We show the RGB image to the user, who clicks on the plug or socket of interest. We then pass the RGB image and pixel coordinates through a powerful segmentation tool \cite{kirillov2023segment} and compute a segmentation mask for the plug or socket. Right: We pass the RGB image, depth image, segmentation mask, and CAD model for the plug or socket into a state-of-the-art pose estimator \cite{wen2023foundationpose} to estimate the 6-DOF pose of the part in the camera frame. We later use robot kinematics and camera extrinsics to convert the pose to the robot frame.}
    \label{fig:perception-pipeline}
\end{figure*}

\begin{figure*}
    \centering
    \includegraphics[width=\textwidth]{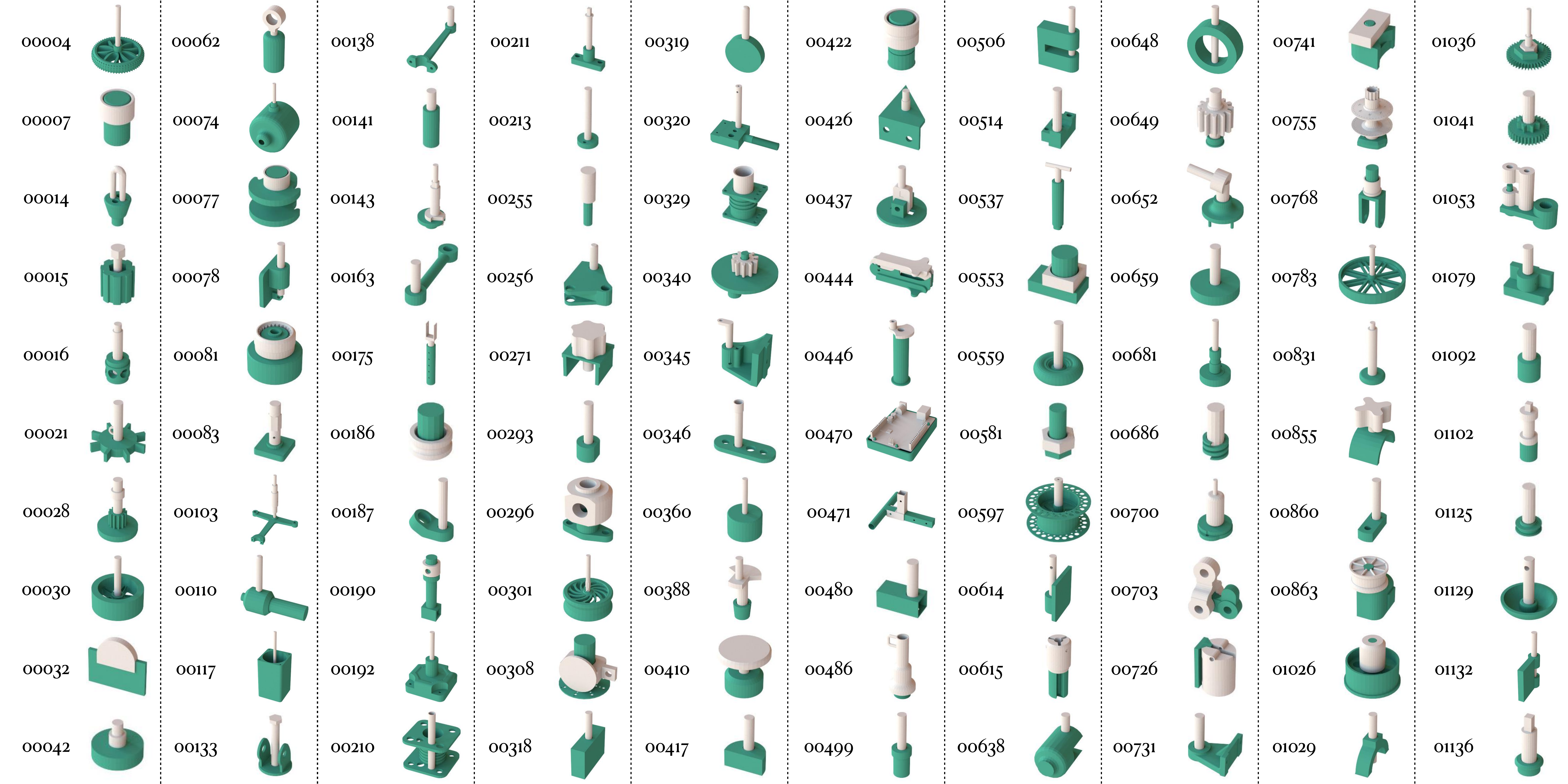}
    \caption{\textbf{Assembly lookup chart.} For each assembly investigated in this work, we provide its unique assembly ID and a rendering. The assemblies are the same as those visualized in \textbf{Figure~\ref{fig:dataset_sim}}. The asset IDs are referenced in figures throughout this paper.}
    \label{fig:asset-lookup}
\end{figure*}

As first described in \textbf{Section~\ref{sec:problem_description}}, our real-world system consists of a Franka robot with a parallel-jaw gripper, a wrist-mounted RealSense D435 camera, a Schunk EGK40 parallel-jaw gripper mounted to the tabletop, and 3D-printed assemblies from our dataset (\textbf{Figure~\ref{fig:real-experimental-setup}}). 
Our communications framework is closely modeled after \cite{tang2023industreal}; 
however, our perception, grasping, and control procedures differ significantly.

For perception, we aim to estimate plug and socket states while initializing them in a far less-constrained manner.
We use a powerful segmentation tool \cite{kirillov2023segment}, textureless CAD models of our parts, and a state-of-the-art pose estimator \cite{wen2023foundationpose} to estimate the 6-DOF poses of each part from RGB-D images. \textbf{Figure~\ref{fig:perception-pipeline}} shows our pipeline; for details, see \textbf{Appendix~\ref{sec:appendix_perception}}.

\subsection{Results: Robustness of Sim-to-Real Transfer to Observation Noise}
\label{sec:appendix_sim_to_real_robustness_obs_noise}

\begin{figure*}
    \centering
    \includegraphics[width=\textwidth]{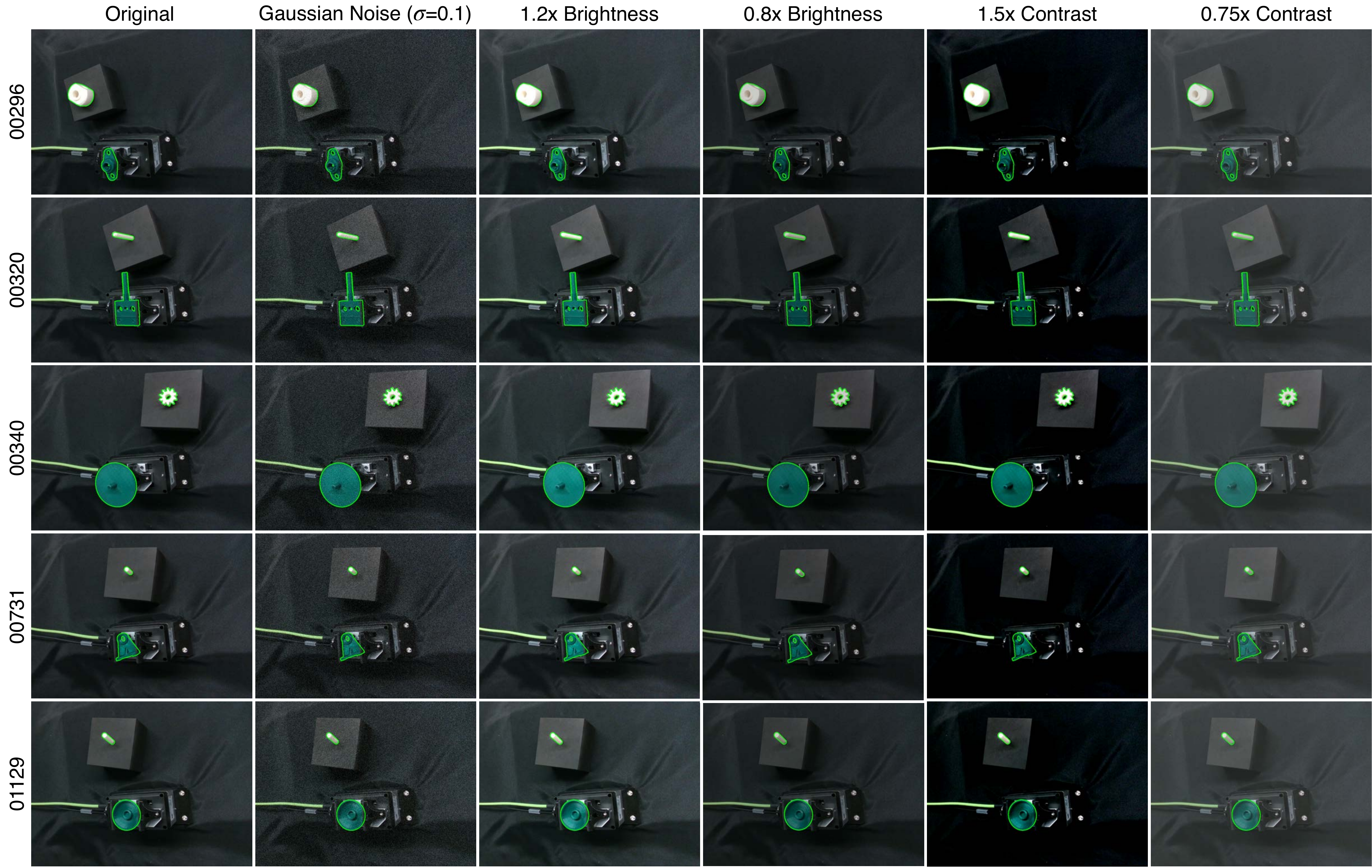}
    \caption{\textbf{Qualitative real-world evaluation of pose estimator with additional image noise.} {For 5 different assemblies, we record camera images, apply 3 types of image augmentation (brightness, contrast, and Gaussian noise), and run the pose estimator. The images show the augmented images, as well as green lines denoting the contours of the plug-and-socket CAD models in the estimated poses. The estimates are accurate and stable across perturbations.}}
    \label{fig:image_noise_w_fp}
\end{figure*}

\textbf{Image noise:} {We train a state-based policy in simulation, use a pose estimator in the real world, and achieve high success rates for the perception-initialized assembly task; thus, our policies are robust to inaccuracies of the pose estimator.

Nevertheless, we evaluate whether our pose estimator would perform well under more adverse conditions.
For 5 different assemblies, we record images, apply 3 types of augmentation (brightness perturbation, contrast perturbation, and Gaussian noise), and run the pose estimator.} \textbf{Figure~\ref{fig:image_noise_w_fp}} {illustrates the results; the estimator is highly robust to image noise.}

\textbf{Control noise:} {We use a torque-controlled robot, implement a task-space impedance controller to generate torques, and achieve high success rates for the assembly task; thus, our policies are robust to errors in our controller.

Nevertheless, we evaluate whether our policies are robust to additional control error.
For 5 different assemblies, we simulate additional error by randomly perturbing the control target by +-2 mm along the $x$, $y$, and $z$ axes and +-5 deg on roll, pitch and yaw at every timestep.
We run 10 trials for each assembly.}
\textbf{Table~\ref{tab:real_control_noise}} {provides our results.
Even with additional control noise, our system maintains high success rates.
We anticipate that performance would degrade at even higher levels of control noise; however, such levels of noise may be unrealistic for our hardware and application.}

\begin{table}[ht]
    \centering
    \begin{tabular}{c|cc}
    \toprule
         & \multicolumn{2}{c}{\textbf{Success Rate}} \\
       Assembly ID & No Control Noise & Control Noise\\\midrule
       \rowcolor{Gainsboro} 00340 & 10/10 & 8/10 \\
        00296 & 8/10 & 9/10 \\
       \rowcolor{Gainsboro} 00731 & 8/10 & 9/10 \\
        01129 & 10/10 & 9/10 \\
       \rowcolor{Gainsboro} 00320 & 10/10& 10/10 \\\bottomrule
     \end{tabular}
    \caption{\textbf{Real-world evaluation of specialist policies with added control noise.} {We deploy our specialist policies for 5 assemblies, with 10 trials per assembly. During each trial, we add +-2 mm of control noise to the $x$, $y$, and $z$ axes, as well as +-5 deg to roll, pitch, and yaw. We compare to deployments under no added noise.}}
    \label{tab:real_control_noise}
\vspace{-20px}
\end{table}

\subsection{Results: Robustness of Sim-to-Real Transfer to Initial-Pose Randomization}
\label{sec:appendix_sim_to_real_robustness_randomization}

\begin{table}[H]
\vspace{-5px}
    \centering
    \begin{tabular}{c|cc}
    \toprule
         & \multicolumn{2}{c}{\textbf{Success Rate}} \\
       Assembly ID & Baseline & 1.5x Baseline \\\midrule
       \rowcolor{Gainsboro} 00340 & 100.0\% (10/10) & 80.0\% (36/45) \\
        00296 & 80.0\% (8/10) & 62.2\% (28/45) \\
       \rowcolor{Gainsboro} 00731 & 80.0\% (8/10) & 73.3\% (33/45) \\
        01129 & 100.0\% (10/10) & 75.6\% (34/45) \\
       \rowcolor{Gainsboro} 00320 & 100.0\% (10/10) & 95.6\% (43/45) \\\bottomrule
     \end{tabular}
    \caption{\textbf{Real-world evaluation of specialist policies with increased initial-pose randomization.} {We deploy our specialist policies for 5 assemblies, with 45 new trials per assembly. We divide the workspace into a 3 $\times$ 3 grid; for each grid cell, we run 5 trials for the corresponding specialist policy while applying 1.5x the initial plug-pose randomization seen during training. We compare to testing over the in-distribution range.}}
    \label{tab:real_plug_rand_automate}
\vspace{-15px}
\end{table}

{We evaluate our specialist policies in the real world with different levels of initial plug- and socket-pose randomization for 5 assemblies.
Specifically, for each assembly, we divide the robot's workspace into a 3 $\times$ 3 grid on the $xy$-plane; for each grid cell, we run
5 trials for the corresponding specialist policy while applying 1.5x the plug-pose randomization seen in training. (See} \textbf{Table~\ref{tab:sim_diff_rand_automate}} {for randomization bounds.)}
\textbf{Table~\ref{tab:real_plug_rand_automate}} {provides the success rates.
Similar to simulated analogues, the policies moderately degrade when out-of-distribution.}

\end{document}